\documentclass[10pt,a4paper]{article}
\usepackage[margin=0.85in]{geometry}
\usepackage{amsmath,amssymb}
\usepackage{booktabs}
\usepackage{graphicx}
\usepackage{xcolor}
\usepackage{hyperref}
\usepackage{array}
\usepackage{caption}
\usepackage{subcaption}
\usepackage{longtable}
\usepackage{float}
\usepackage{placeins}
\usepackage{multirow}
\hypersetup{colorlinks=true,linkcolor=blue,citecolor=blue,urlcolor=blue}
\graphicspath{{figures/}}

\title{Curriculum-Based Deep Reinforcement Learning for Robust Autonomous Rocket Landing under Atmospheric Uncertainty}
\author{
\begin{tabular}{cc}
\begin{tabular}{c}
Atin Chowdhury\\[3pt]
\href{mailto:atin.c23@iiits.in}{\small atin.c23@iiits.in}
\end{tabular}
&
\begin{tabular}{c}
Dr. Piyush Joshi\\[3pt]
\href{mailto:piyush.j@iiits.in}{\small piyush.j@iiits.in}
\end{tabular}
\end{tabular}
\\[0.8em]
\small Department of Computer Science and Engineering\\
\small Indian Institute of Information Technology, Sri City, Andhra Pradesh, India
}
\date{}

\begin{document}
\maketitle

\begin{abstract}
Autonomous far-range rocket landing is a challenging control problem because navigation, descent-rate regulation, attitude management, and fuel use must be coordinated in the presence of uncertain atmospheric and vehicle dynamics. This paper presents a curriculum-driven proximal policy optimization (PPO) framework for three-dimensional rocket landing in a stochastic environment incorporating altitude-dependent aerodynamic drag, wind shear, temporally correlated turbulence, discrete gusts, uncertain propulsion and aerodynamic parameters, thrust-vector misalignment, engine lag, actuator-gain variation, noisy observations, and command latency. Learning is organised into far-range approach, soft descent, and integrated landing stages. Policy transfer, critic warm-up after reward changes, rehearsal of earlier tasks, and progressive expansion of weather severity are used to support learning across these coupled phases. Across 100,000 evaluation episodes, the policy achieves 96,656 successful landings, corresponding to a success rate of 96.656\% and a 95\% Wilson confidence interval of 96.543--96.766\%. Success decreases from 98.84\% under zero-disturbance conditions to 88.07\% in the highest weather-severity band. Successful touchdowns exhibit small miss distances and low residual horizontal and vertical velocities. An exploratory logistic-regression analysis identifies mean wind speed as the strongest measured correlate of success, with initial distance, initial altitude, initial vertical velocity, and wind-shear exponent also associated with outcomes. The results quantify both the effectiveness of the learned policy over the evaluated disturbance distribution and the degradation that occurs under severe compound atmospheric conditions. Landing visualizations are available in \href{https://drive.google.com/file/d/1ieIX0FJW_Cux6n7KkX9IKadc2h-7Lo4m/view?usp=drive_link}{Video 1} and \href{https://drive.google.com/file/d/1Ng0jM4dUlp52AFoMK9HlYBP-6-AiLjOe/view?usp=drive_link}{Video 2}.
\end{abstract}


\section{Introduction}

Reusable launch vehicles require the terminal stage of flight to bring a vehicle from a large initial displacement to a small landing region while regulating vertical and horizontal velocity, attitude, angular rate, and fuel consumption. The difficulty is compounded by the coupling between translation and rotation: the direction of thrust depends on the vehicle's attitude, while the attitude must often be changed to first accelerate laterally and then counter-thrust to remove the velocity accumulated during approach. The resulting problem is nonlinear, hybrid, and constrained. A command that is locally useful for reducing horizontal distance can be harmful if the vehicle has already exceeded a braking-speed envelope or is descending too quickly to recover before touchdown.

Operational launch systems rely on engineered guidance, state estimation, actuator design, and trajectory optimisation. Convex and successive-convexification methods have provided structured approaches to powered descent under dynamical and path constraints~\cite{acikmese2013lossless,blackmore2010minimum}; broader discussions of precision rocket landing highlight the need to couple trajectory planning with reliable control and state estimation~\cite{blackmore2016}. Deep reinforcement learning (RL) offers a complementary route: a policy can be trained by repeated simulation interaction and then evaluated as a feedback controller, avoiding the need to hand-code every manoeuvre. RL has been investigated for six-degree-of-freedom planetary landing and terminal guidance~\cite{gaudet2020deep,gaudet2020terminal,gaudet2021mars}, as well as for reusable-rocket landing in custom environments~\cite{chowdhury2024icaart,rathod2023,su2022}.

However, landing from a distant horizontal position is not simply a vertical-descent problem with a larger initial offset. The controller must determine when to accelerate toward the pad, when to reverse thrust to brake lateral motion, and how to coordinate those manoeuvres with vertical descent. Under atmospheric uncertainty, the effective input-output relationship changes over time: the same command can yield a different acceleration because of wind-relative drag, thrust variation, engine lag, or command delay. Sensor perturbations further mean that the policy does not have exact access to the physical state. Training only under nominal conditions can therefore produce a policy that succeeds on a narrow set of trajectories but degrades when disturbances change.

This paper presents a curriculum-driven PPO controller for three-dimensional far-range rocket landing with an explicit stochastic atmosphere and vehicle-uncertainty model. The simulator includes altitude-dependent quadratic drag, wind shear and slow wind drift, correlated turbulent fluctuations, finite-duration gusts, uncertain air density and drag, uncertain thrust, thrust-vector misalignment, fuel-dependent acceleration, engine spool-up lag, control-gain variation, noisy measurements, and command latency. The atmosphere's latent state is withheld from the policy; the agent must infer the need for corrective actions from the available noisy observations and its previous interaction with the environment. During training, a weather curriculum gradually raises the maximum disturbance level while retaining a fraction of calm episodes.

The central research question is not whether a simulated policy can land under one nominal configuration, but how its landing probability and touchdown quality vary over a prescribed distribution of atmospheric and model perturbations. Accordingly, we analyse the aggregate outcome of 100,000 evaluation episodes, stratify success by weather severity and selected measured disturbance variables, inspect landing quality conditional on success, and discuss the failure modes that remain. 

The paper makes four contributions. First, it documents a parameterized 3D landing environment in which atmospheric forces, wind, turbulence, gusts, sensor noise, actuator uncertainty, and engine response uncertainty affect the vehicle dynamics. Second, it formulates a three-stage curriculum with policy transfer, critic warm-up following a reward-mode change, rehearsal of earlier skills, and a progressive weather-severity schedule. Third, it gives a mathematical account of the stage-dependent reward terms, including braking envelopes, radial progress, directional thrust alignment, touchdown shaping, and terminal adjustments. Fourth, it reports a large evaluation with severity-conditioned outcomes and exploratory associations between success and disturbance variables, making performance degradation visible rather than hiding it behind a single average. The remainder of the paper describes the related literature, formalises the simulator and learning problem, presents the reward and training methodology, and analyses the evaluation results and limitations.

\section{Related Work}

Research on autonomous rocket landing spans model-based guidance, deep reinforcement learning, curriculum design, and robustness to uncertainty. These strands address related but distinct questions. Optimisation-based guidance provides principled ways to satisfy path and terminal constraints, whereas learning-based control seeks a policy that can map observations to actions with low inference-time cost. Robustness research asks a further question: whether the resulting controller remains effective when the dynamics, measurements, or disturbances differ from those encountered during nominal training.

Lossless convexification and successive convexification have been used to reformulate challenging powered-descent problems into tractable optimisation problems while respecting constraints on thrust and trajectory~\cite{acikmese2013lossless,blackmore2010minimum}. These methods are attractive because their optimisation structure can expose feasibility and constraint trade-offs. Their performance, however, depends on the fidelity of the model, the formulation of constraints, and the computational procedure used to solve the guidance problem. General discussions of precision rocket landing also stress that guidance must be integrated with state estimation and vehicle control rather than treated as a purely geometric targeting problem~\cite{blackmore2016}. The present study takes a learning-based route and focuses on the effect of stochastic atmospheric and vehicle perturbations within a custom simulator; it does not claim that PPO replaces validated flight guidance or that the simplified simulator reproduces a flight-qualified vehicle model.
Gaudet, Linares, and Furfaro investigated deep RL for six-degree-of-freedom planetary landing and terminal adaptive guidance~\cite{gaudet2020deep,gaudet2020terminal,gaudet2021mars}. This line of work demonstrates the relevance of learned feedback policies to nonlinear descent tasks. For reusable rocket landing, Chowdhury et al.~\cite{chowdhury2024icaart} introduced a custom 3D simulation and compared PPO, DQN, and Double DQN; further simulation studies have considered curriculum or imitation learning, jump-start mechanisms, and shaped landing rewards~\cite{rathod2023,jiang2024,su2022}. More recent developments include adaptive PPO for precision landing, thrust-vectored guidance, hierarchical learned guidance with a lower-level controller, and hardware-aware acceleration~\cite{zhu2025adaptppo,ieee2022thrust,ieee2025hierarchical,nature2026fpga}. These approaches differ in dynamics, action spaces, training assistance, and evaluation protocol, so their headline success rates should not be compared without accounting for initial-state distributions, touchdown criteria, and test conditions.

A curriculum breaks a difficult learning problem into a sequence of tasks or distributions, enabling the policy to learn useful behaviour before the full task is introduced. Narvekar et al.~\cite{narvekar2020curriculum} survey curriculum design and distinguish aspects such as task decomposition, task ordering, and transfer mechanism. In this work, the curriculum is defined around three physical subproblems: far-range approach, soft descent near the pad, and an integrated landing task. The policy parameters are transferred between stages. When the reward mode changes, the critic is reinitialised and the actor is temporarily frozen so that early value-estimation errors do not immediately destabilise the transferred policy. Rehearsal episodes from earlier stages are mixed into later stages to preserve previously learned behaviour. This transfer protocol is intended to reduce interference between skills, although controlled ablations are required to quantify the independent contribution of each component.

Replay and rehearsal methods have been studied as a means of stabilising learning across changes in the data distribution; prioritized experience replay is a representative example of using stored transitions to change the sampling distribution during learning~\cite{schaul2015prioritized}. The proposed method uses task-level rehearsal: it samples complete episodes from earlier curriculum stages rather than prioritising a fixed transition buffer. Separately, the atmospheric curriculum randomises the severity of wind, turbulence, gusts, sensing error, propulsion error, and actuator response. This is related in spirit to training over a distribution of environment parameters rather than a single nominal system, but the current experiments do not isolate the effect of each source of uncertainty in controlled ablations. Therefore, the results below are best interpreted as an empirical robustness profile over the chosen joint perturbation distribution. They do not establish causal contributions of individual disturbance variables or a sim-to-real transfer guarantee.

In summary, the distinction of the present study is the combination of a three-stage far-range landing curriculum with a progressively widened atmosphere/vehicle uncertainty distribution and a 100,000-episode evaluation. The focus is the measured dependence of success and touchdown quality on weather severity and selected disturbance variables, together with an explicit account of the simplified simulator and the remaining validation requirements.

\section{Environment and Problem Formulation}
\label{sec:environment}

This section formalises the vehicle state, action interface, per-frame simulation update, stochastic atmosphere, measurement process, and touchdown task. The simulator is a reduced-order environment designed for repeatable reinforcement-learning experiments; it is not a validated aerodynamic model of a particular launch vehicle. We distinguish the internal simulation state from the observation supplied to the policy: the dynamics evolve using the latent state and disturbances, while the controller receives a scaled and potentially noisy 29-dimensional observation. This distinction is important because the policy must react to atmospheric effects without directly observing the wind or all randomised vehicle parameters.

The simulation advances at 60 physics frames per second. One agent decision spans three physics frames, so a policy action is held for the decision interval except for the initial portion of that interval when command latency requires the previous command to remain active. Unless otherwise stated, positions, velocities, forces, and distances are reported in the simulator's native units; velocities and per-frame acceleration coefficients use units/frame and units/frame$^2$, respectively.

\subsection{State, actions, and discrete-time dynamics}

Let the latent vehicle state at physics frame $k$ be represented by
\begin{equation}
\mathbf{x}_k = (\mathbf{p}_k,\mathbf{v}_k,R_k,\boldsymbol{\omega}_k,F_k,q_k),
\end{equation}
where $\mathbf{p}_k=[x_k,y_k,z_k]^\top$ is position, $\mathbf{v}_k=[v_{x,k},v_{y,k},v_{z,k}]^\top$ is velocity, $R_k\in SO(3)$ is the attitude matrix, $\boldsymbol{\omega}_k$ collects the pitch and roll rates, $F_k\in[0,F_{\max}]$ is remaining fuel, and $q_k$ is the internal engine-thrust state. The augmented latent state also includes atmospheric states (wind drift, turbulence and gust state), sampled episode parameters, sensor-noise state, and the previous action required to implement latency. The observation omits parts of this augmented state, so the agent faces a partially observable control problem even though the simulator's augmented transition is Markovian.

The policy acts in the product action space
\begin{equation}
 a_t=(u_t^{\mathrm{thr}},u_t^{\mathrm{pitch}},u_t^{\mathrm{roll}})
 \in\{0,1\}\times\{0,1,2\}\times\{0,1,2\}.
\end{equation}
Here $u^{\mathrm{thr}}=1$ requests thrust; each attitude command selects no rotation or one of two opposite tilt directions. The space contains $2\cdot3\cdot3=18$ joint command tuples. In the implementation, the PPO policy uses the MultiDiscrete action interface, represented as categorical factors for the thrust, pitch, and roll components.

The rotation matrix stores the vehicle axes in world coordinates, with the nose unit vector $\mathbf{n}_k$ identifying the nominal thrust direction. In compact form the attitude integration is
\begin{equation}
 R_{k+1}=\operatorname{Renorm}\!\left(R_k R_x(\omega_{P,k})R_z(\omega_{R,k})\right),\qquad
 \boldsymbol{\omega}_{k+1}=0.985\left[\boldsymbol{\omega}_{k}+\Delta\boldsymbol{\omega}_{k}^{\mathrm{cmd}}+\Delta\boldsymbol{\omega}_{k}^{\mathrm{aero}}\right],
\end{equation}
where the rotation update follows the vehicle's body axes and the matrix is periodically renormalised for numerical stability. The command-induced rate update depends on current tilt, thrust state, and actuator gain; the atmospheric model can also add an aerodynamic weathercock torque. This equation is schematic: the exact ordered frame updates and clamping are implemented in \,\texttt{RocketSim.frame} and are not replaced by a continuous-time rigid-body integrator.

At each physics frame, thrust is applied along a perturbed nose vector. If $q_k\in[0,1]$ denotes engine spool state, $s_T$ the episode thrust scale, and $m(F_k)$ the simplified fuel-dependent gain, the scalar thrust coefficient is
\begin{equation}
 T_k=T_0q_ks_T\left[1+\mu\left(1-\frac{F_k}{F_{\max}}\right)\right],\quad T_0=0.02,\quad \mu=0.15.
\end{equation}
The spool state tracks the binary thrust request through
\begin{equation}
 q_{k+1}=q_k+\alpha_q(u_k^{\mathrm{thr}}-q_k),
\end{equation}
where $\alpha_q=1$ in the no-lag case and otherwise depends on a sampled lag time constant. Thrust-vector misalignment perturbs the commanded direction by small components along the two body axes. The velocity update includes this thrust contribution, gravity $g=-0.007$ units/frame$^2$, and atmospheric drag; position is then advanced using the updated velocity. Fuel is finite, with $F_{\max}=1000$ units. Thrust consumes $12/60$ fuel units per frame while active, and an attitude key consumes $2.5/60$ units per frame while held.

\subsection{Atmospheric forces and wind field}

The atmosphere is parameterised by a scalar severity level $\ell\in[0,1]$. At level zero, the random wind, turbulence, gust, and uncertainty magnitudes vanish, but deterministic air density and air resistance remain active when the atmospheric branch is enabled; therefore, “calm” means zero stochastic weather severity, not vacuum. At positive levels, the random wind, turbulence, gusts, uncertain vehicle parameters, and measurement perturbations are active as well. Air density is represented by an exponential profile
\begin{equation}
 \rho_k=s_\rho\exp\!\left(-\frac{\max(z_k,0)}{H_\rho}\right),\qquad H_\rho=3000,
\end{equation}
where $s_\rho$ is an uncertain scale factor. The wind vector is $\mathbf{w}_k=[w_{x,k},w_{y,k},w_{z,k}]^\top$, and drag depends on air-relative velocity
\begin{equation}
 \mathbf{v}^{\mathrm{rel}}_k=\mathbf{v}_k-\mathbf{w}_k,\quad V^{\mathrm{rel}}_k=\|\mathbf{v}^{\mathrm{rel}}_k\|,\quad v_{\parallel,k}=\mathbf{v}^{\mathrm{rel}}_k\cdot\mathbf{n}_k,\quad
 \mathbf{v}_{\perp,k}=\mathbf{v}^{\mathrm{rel}}_k-v_{\parallel,k}\mathbf{n}_k.
\end{equation}
The per-frame drag update is anisotropic:
\begin{equation}
 \Delta\mathbf{v}_{D,k}=-\rho_k s_D V^{\mathrm{rel}}_k\left(K_{\parallel}v_{\parallel,k}\mathbf{n}_k+K_{\perp}\mathbf{v}_{\perp,k}\right),
 \quad K_{\parallel}=4\times10^{-4},\quad K_{\perp}=1.6\times10^{-3},
\end{equation}
with uncertain drag scale $s_D$. Thus the broadside coefficient is four times the nose-axis coefficient. The model also adds a small attitude-rate torque proportional to air-relative speed, density, and the lateral projections of the relative-wind vector. It represents a simplified aerodynamic tendency to weathercock and is not a full aerodynamic moment model.

The mean wind has a random horizontal direction $\phi$ and a power-law vertical shear profile. Let $z_r=300$ and $\alpha_s\in[0.10,0.30]$ be the reference altitude and sampled shear exponent. The shear multiplier is
\begin{equation}
 S(z)=\min\!\left[\left(\frac{\max(z,z_{\min})}{z_r}\right)^{\alpha_s},2\right],\qquad z_{\min}=10.
\end{equation}
Then the horizontal wind combines the height-scaled mean vector, a slowly varying drift, turbulence, and any active gust; the vertical wind similarly combines a sampled mean component and vertical turbulence/gust. Turbulence is updated as a correlated autoregressive process:
\begin{equation}
 u_{k+1}=a u_k+\sigma_\ell\sqrt{1-a^2}\,\epsilon_k,\qquad
 a=\exp(-1/\tau),\quad \epsilon_k\sim\mathcal{N}(0,1),
\end{equation}
where $\tau\sim U(40,100)$ frames and $\sigma_\ell$ is the episode's severity-scaled horizontal turbulence standard deviation. The vertical component uses half the innovation scale. A ground-amplification factor increases turbulence close to the surface. Discrete gusts begin according to a per-frame Bernoulli event with probability $p_g$ and use a finite-duration one-cosine envelope,
\begin{equation}
 G_k=\frac{1}{2}\left[1-\cos\!\left(\frac{2\pi j}{D_g}\right)\right],\qquad j=0,\ldots,D_g-1,
\end{equation}
where the sampled duration $D_g$ is between 60 and 160 frames. These definitions produce temporally correlated disturbances rather than independent white-noise wind at each frame.

\subsection{Episode randomisation and observation model}

The weather sampler draws a level of zero with probability $0.15$. Otherwise the level is uniform on $[0,\ell_{\max}]$, where $\ell_{\max}$ is the current weather-curriculum cap. Conditional on $\ell$, the implementation samples the principal uncertain parameters from the ranges in Table~\ref{tab:uncertainty}. These random variables are sampled per episode and remain fixed unless described as a time-varying process. In particular, thrust, density, drag, tilt-control authority, engine response, thrust misalignment, and command latency are not independently observed by the policy.

\begin{table}[H]
\centering
\small
\caption{Principal uncertainty ranges in the stochastic environment. Ranges describe the simulator's engineering model and are not flight-calibrated distributions.}
\label{tab:uncertainty}
\begin{tabular}{p{4.1cm}p{3.7cm}p{5.5cm}}
\toprule
Quantity & Range at $\ell=1$ & Implementation role\\
\midrule
Reference horizontal wind & $[0.4,1.0]\times1.5$ units/frame & Random direction and altitude shear\\
Horizontal turbulence SD & $[0.4,1.0]\times0.25$ units/frame & Correlated per-frame perturbation\\
Vertical turbulence innovation & Half the horizontal scale & Vertical atmospheric fluctuation\\
Gust amplitude parameter & $[0.4,0.9]\times1.5$ units/frame & Intermittent 60--160-frame gusts\\
Gust initiation probability & $[0,0.0015]$ per frame & Probability of starting a gust\\
Density scale $s_\rho$ & $[0.85,1.15]$ & Scales air density\\
Drag scale $s_D$ & $[0.75,1.25]$ & Scales drag coefficients\\
Thrust scale $s_T$ & $[0.94,1.06]$ & Scales thrust magnitude\\
Thrust-vector misalignment & $\pm0.6^\circ$ per axis & Perturbs effective thrust direction\\
Tilt-control gain & $[0.90,1.10]$ & Changes control authority\\
Engine lag time constant & $[0,2.5]$ frames & Smooths thrust start/stop response\\
Command latency & 0--2 frames & Retains previous action temporarily\\
Sensor noise & Severity-dependent & Position, altitude, velocity, attitude, and rate estimates\\
\bottomrule
\end{tabular}
\end{table}

The policy observation is $\mathbf{o}_t\in[-5,5]^{29}$. It includes vertical components of the body axes, normalised pitch and roll rates, tilt-command state, vertical velocity, altitude above the pad, fuel fraction, stopping margins, horizontal coordinates and velocity, body-frame projections of horizontal position and velocity, nose horizontal components, hover-height error, radial distance, radial velocity, and fine-scale position/velocity features. No wind vector is exposed. For $\ell>0$, the sensor model perturbs horizontal position with a standard deviation that grows with range, altitude, velocity, angular-rate, and attitude measurements with severity-dependent noise. Hence the observation may be written abstractly as
\begin{equation}
 \mathbf{o}_t=\operatorname{clip}_{[-5,5]}\!\left(g(\mathbf{x}_{3t},\boldsymbol{\nu}_{t};\boldsymbol{\eta})\right),
\end{equation}
where $\boldsymbol{\eta}$ denotes the fixed per-episode uncertain parameters and $\boldsymbol{\nu}_t$ the measurement perturbations. The reward and touchdown checks use the underlying simulator state, not the noisy observation. The setup is consequently a partially observed stochastic control problem. The implemented policy is feed-forward: it maps the current noisy observation to an action and does not maintain an explicit recurrent state or belief over the hidden weather and vehicle parameters.

\subsection{Landing task, termination, and evaluation criterion}

The three curriculum tasks share the same three-dimensional vehicle model but use different initial-state distributions and reward weights. Stage 1 begins 1,000--1,500 units horizontally from the pad and targets a hover region at altitude $z=300$. Stage 2 begins near the pad with a faster descent and focuses on soft touchdown. Stage 3 combines the far-range approach with terminal descent. Attitude remains part of the observed and rewarded state, but Stage 3 does not impose a separate upright-touchdown requirement.

Landing success is defined by three touchdown conditions: horizontal distance from the pad centre no greater than $R_{\mathrm{pad}}=16$ simulator units, vertical velocity $v_z\geq-0.2$ units/frame, and horizontal speed $v_h\leq0.25$ units/frame. Stage 3 does not impose a separate uprightness threshold; tilt is reported as a descriptive touchdown quantity. The reward retains tilt-dependent shaping in both the touchdown bonus and the failed-touchdown penalty, but these reward terms do not add a tilt predicate to the Boolean success label.

The control objective is to maximise expected discounted return under the joint distribution of initial states, weather parameters, and time-varying disturbances while satisfying touchdown constraints as frequently as possible. If $Y\in\{0,1\}$ denotes the binary landing outcome, the reported aggregate metric is $\hat p=N^{-1}\sum_{i=1}^N Y_i$ for $N=100{,}000$ episodes. This empirical quantity estimates success probability under the specified evaluation distribution; it does not establish success probability outside that distribution.

\section{Proposed Architecture}
\label{sec:architecture}
The landing problem couples perception, decision-making, actuation, and vehicle motion in a feedback process. At each decision instant, the policy must infer from imperfect measurements whether the vehicle should build lateral velocity, reverse thrust to brake, reduce its descent rate, or prioritise terminal alignment. These objectives interact: a correction that reduces horizontal miss distance can increase lateral velocity, while a late attitude change can leave insufficient time to arrest vertical motion. The architecture is therefore organised around repeated interaction between the learned policy and a flight environment in which atmospheric conditions, vehicle parameters, sensor measurements, and actuator response may vary.

\begin{figure}[H]
\centering
\includegraphics[width=\textwidth,height=0.50\textheight,keepaspectratio]{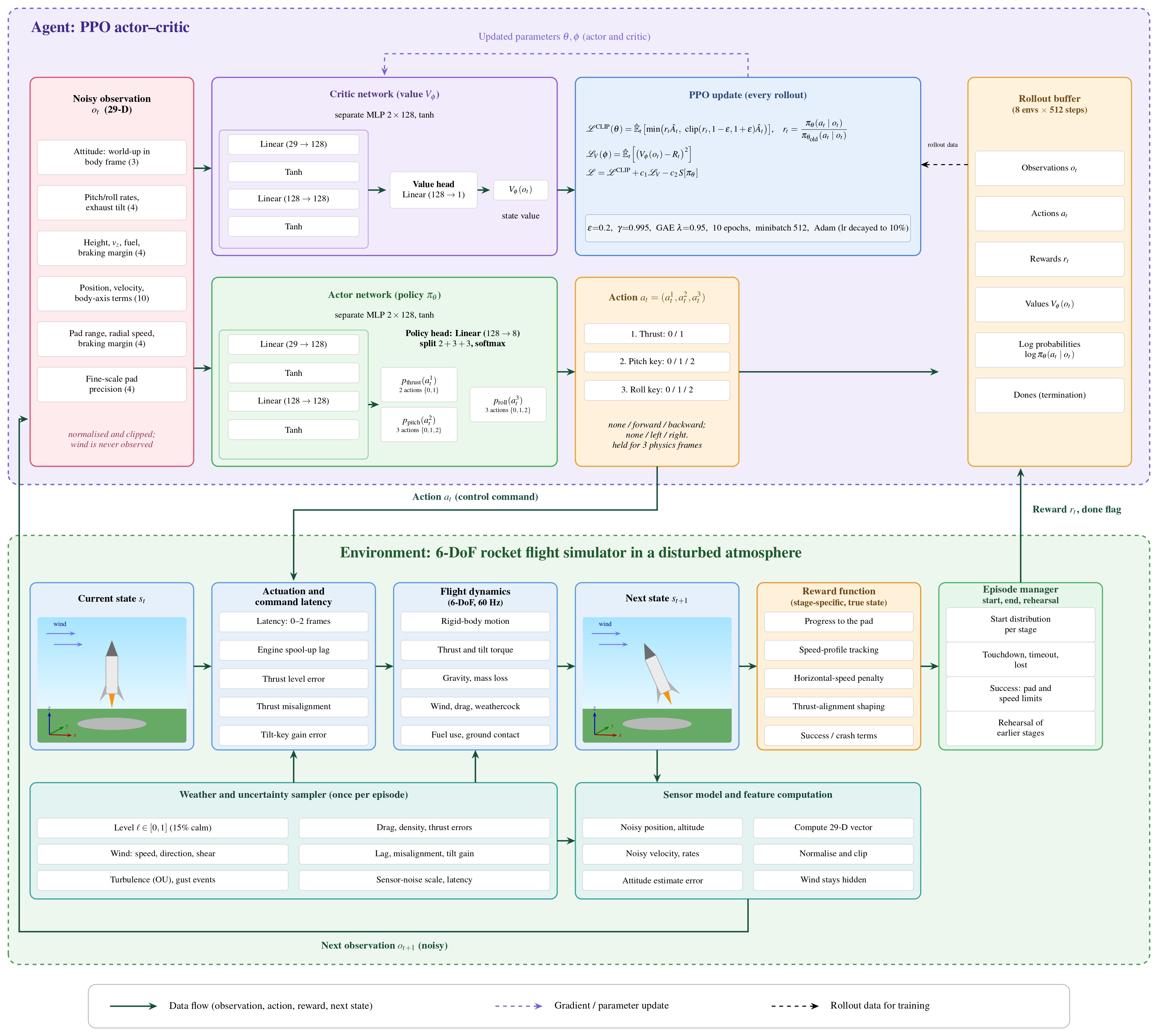}
\caption{Overall architecture of the curriculum-trained PPO rocket-landing system. The diagram shows the 29-dimensional noisy observation, separate actor and critic networks, factorised MultiDiscrete action head, rollout buffer, PPO update, disturbance sampler, actuator latency, flight dynamics, reward calculation, and episode manager. Although the schematic includes the label ``6-DoF'', the simulation model used in this study approximates three-dimensional translation and commanded pitch/roll attitude dynamics; it is not a validated full six-degree-of-freedom aerodynamic vehicle model.}
\label{fig:architecture}
\end{figure}

Figure~\ref{fig:architecture} presents the relationship between the policy-learning components and the flight environment. The environment converts the latent state into a normalised observation, the actor selects thrust, pitch, and roll commands, and the actuator model introduces the delay and response dynamics that determine the effective input. The vehicle then evolves under gravity, propulsion, aerodynamic forces, and atmospheric disturbances. A stage-dependent reward evaluates the transition, while the episode manager applies the active initial-state distribution and termination rules. Each transition is stored for subsequent PPO optimisation, connecting the faster flight-control loop to the slower parameter-update loop. The wind field and episode-level hardware perturbations are not directly exposed to the actor, so disturbance rejection must emerge from feedback based on the available observations.

\FloatBarrier

\subsection{Actor--critic policy and rollout optimisation}

The policy receives $\mathbf{o}_t\in[-5,5]^{29}$ and processes it using a multilayer perceptron with two hidden layers of 128 units and $\tanh$ activations. The actor and critic use separate network branches. The actor's linear output layer produces eight logits, which are split into groups of sizes two, three, and three for the thrust, pitch, and roll choices. Softmax is applied within each group, defining the factorised action distribution
\begin{equation}
\pi_{\theta}(a_t\mid\mathbf{o}_t)=
\pi_{\theta}^{\mathrm{thr}}(u_t^{\mathrm{thr}}\mid\mathbf{o}_t)
\pi_{\theta}^{\mathrm{pitch}}(u_t^{\mathrm{pitch}}\mid\mathbf{o}_t)
\pi_{\theta}^{\mathrm{roll}}(u_t^{\mathrm{roll}}\mid\mathbf{o}_t).
\end{equation}
The resulting action contains $2\times3\times3=18$ possible joint commands. The critic branch has the same two-layer hidden structure and ends in a scalar value estimate $V_{\phi}(\mathbf{o}_t)$, which approximates expected discounted return from the current observation under the policy and training distribution.

Eight parallel environments each contribute 512 agent decisions per rollout, yielding approximately $8\times512=4096$ transitions before an update. Each transition stores the observation, action, reward, estimated value, log probability, and episode-boundary information. PPO forms the probability ratio
\begin{equation}
 r_t(\theta)=\exp\!\left[\log\pi_{\theta}(a_t\mid\mathbf{o}_t)
 -\log\pi_{\theta_{\mathrm{old}}}(a_t\mid\mathbf{o}_t)\right]
\end{equation}
and optimises a clipped surrogate objective using the generalised advantage estimate $\widehat A_t$. The clipping interval is $[1-\epsilon,1+\epsilon]$ with $\epsilon=0.2$. The objective also includes a value-prediction loss and an entropy term. We use $\gamma=0.995$, GAE parameter $\lambda=0.95$, 10 optimisation epochs per rollout, minibatches of 512, and Adam with a learning rate decaying linearly to 10\% of its initial stage-specific value. The actor and critic are therefore updated from the same sampled rollout but are trained to solve distinct problems: selecting actions and estimating returns, respectively.

\subsection{Environment feedback and disturbance injection}

On the environment side, each policy command is interpreted by a discrete actuator model that includes a sampled command latency of zero to two physics frames, engine spool-up lag, uncertain thrust magnitude, thrust-vector misalignment, and variation in tilt-control gain. The simplified flight update combines thrust along the perturbed nose direction, gravity, air-relative drag, fuel consumption, and the implemented attitude-rate update. Atmospheric effects include an altitude-dependent density profile, a sheared mean wind, correlated turbulence, and finite-duration gusts. Some uncertain parameters are sampled once per episode, while wind drift, turbulence, and gust envelopes evolve during the episode.

The sensor and feature-computation block converts the latent state into the clipped observation vector used by the actor and critic. Noise affects position, altitude, velocity, attitude, and rate measurements; the wind state and episode-level hardware perturbations are not directly exposed. The reward module uses the true simulator state to calculate progress, speed-profile penalties, thrust-alignment shaping, and terminal adjustments. The episode manager applies the current stage's initial-state distribution and termination conditions, and may sample rehearsal episodes from earlier stages. This separation between latent dynamics, noisy observations, reward calculation, and PPO rollout data is central to interpreting the setup as a partially observed stochastic control problem.

\FloatBarrier

\section{Reward Function: Mathematical Formulation and Rationale}
\label{sec:reward}

The reward is designed to make the long-range landing objective learnable without prescribing a fixed trajectory. It combines local penalties for unsafe speed or attitude with a signed progress term and directional cues for thrust use. The terms are stage-dependent because navigation, soft descent, and the integrated task have different immediate learning needs. A common scale factor of $0.1$ is applied to the raw reward before returning it to PPO. In the equations below, $[x]_+=\max(x,0)$ and $\operatorname{clip}(x;a,b)=\min(\max(x,a),b)$.

The reward design has three intended roles. First, progress shaping should distinguish actions that reduce the pad-relative distance from those that move away. Second, braking envelopes should discourage horizontal or vertical speeds that are inconsistent with the remaining distance and height. Third, local and terminal terms should refine touchdown while penalising excessive angular motion, thrust usage, and time. The directional alignment terms provide a soft preference for pointing thrust toward the pad when the vehicle is below its horizontal speed envelope and against the current motion when it is above that envelope. They do not prescribe a scripted sequence: the policy still selects thrust, pitch, and roll jointly at every decision.

\subsection{Shared geometric quantities and directional alignment}

Let the horizontal distance and horizontal speed at agent decision $t$ be
\begin{equation}
 d_t=\sqrt{x_t^2+y_t^2},\qquad v_{h,t}=\sqrt{v_{x,t}^2+v_{y,t}^2},\qquad h_t=\max(z_t-z_{\mathrm{pad}},0),
\end{equation}
where $z_{\mathrm{pad}}=13$. The horizontal braking envelope is
\begin{equation}
 v_{\mathrm{cap},t}=\sqrt{2A_Hd_t}+0.05,\qquad A_H=0.004.
\end{equation}
It follows from the constant-deceleration stopping relation $v^2=2ad$, using a conservative design deceleration $A_H$; the additive $0.05$ is a small speed allowance. This is a shaping reference, not a hard state constraint. The horizontal overspeed feature is
\begin{equation}
 q_{h,t}=\operatorname{clip}\!\left(\frac{[v_{h,t}-v_{\mathrm{cap},t}]_+}{0.3};0,2\right).
\end{equation}

Write $n_h=\sqrt{n_x^2+n_y^2}$ for the magnitude of the horizontal projection of the nose. When the associated vectors are nonzero, define the cosine-like directional terms
\begin{equation}
 c_{d,t}=-\frac{x_tn_x+y_tn_y}{d_tn_h},\qquad
 c_{v,t}=-\frac{v_{x,t}n_x+v_{y,t}n_y}{v_{h,t}n_h}.
\end{equation}
The first is positive when the nose's horizontal projection points toward the pad; the second is positive when it points against horizontal velocity. The need for acceleration and braking is modulated continuously by
\begin{equation}
 a_{\mathrm{need},t}=\operatorname{clip}\!\left(\frac{v_{\mathrm{cap},t}-v_{h,t}}{3};0,1\right),\qquad
 b_{\mathrm{need},t}=\operatorname{clip}\!\left(\frac{v_{h,t}-v_{\mathrm{cap},t}}{2};0,1\right).
\end{equation}
For thrust command $u_t^{\mathrm{thr}}\in\{0,1\}$, the alignment reward is
\begin{equation}
 r_{\mathrm{align},t}=w_a u_t^{\mathrm{thr}}n_h[c_{d,t}]_+a_{\mathrm{need},t}
 +w_b u_t^{\mathrm{thr}}n_h[c_{v,t}]_+b_{\mathrm{need},t},
\end{equation}
with a zero value assigned when a normalising vector is too small. For stage three, $w_a=0.45$ and $w_b=0.90$; the corresponding stage-one weights are $0.70$ and $1.00$. Thus the policy is rewarded for thrust aligned toward the pad when additional lateral speed is useful and for counter-thrust when the approach speed exceeds the envelope.

\subsection{Stage-one navigation reward}

Stage one must reach and settle inside the hover region around $(x,y,z)=(0,0,300)$. Let $\Delta z_t=z_t-z_{\mathrm{hov}}$ and $z_{\mathrm{hov}}=300$. A signed vertical reference velocity is defined as
\begin{equation}
 v^{(1)}_{\mathrm{des},t}=\begin{cases}
 -\sqrt{2A_{\mathrm{des}}\Delta z_t},&\Delta z_t>0,\\
 +\sqrt{2(0.005)(-\Delta z_t)},&\Delta z_t\leq0,
 \end{cases}\qquad A_{\mathrm{des}}=0.0078.
\end{equation}
The associated vertical overspeed indicator is normalised and saturated at two. Let $q_{z,t}=\operatorname{clip}(e_{z,t}/(|v^{(1)}_{\mathrm{des},t}|+0.2);0,2)$, where $e_{z,t}=[v^{(1)}_{\mathrm{des},t}-v_{z,t}]_+$ above the hover altitude and $e_{z,t}=[v_{z,t}-v^{(1)}_{\mathrm{des},t}]_+$ below it. Let $\theta_t$ be tilt from vertical and $\omega_t=\sqrt{\omega_{P,t}^2+\omega_{R,t}^2}$ the angular-rate norm. The raw stage-one reward is
\begin{align}
 R^{(1)}_t={}&w_{\mathrm{prog}}(d_{t-1}-d_t)
 +w_{\mathrm{prog},z}(|\Delta z_{t-1}|-|\Delta z_t|)
 -w_{\mathrm{ov}}q_{h,t}-w_{\mathrm{ovz}}q_{z,t}\nonumber\\
 &-w_{\mathrm{orient}}\left(\frac{\theta_t}{90^\circ}\right)^2
 -w_\omega\omega_t-w_{vh}\min(v_{h,t},1)e^{-d_t/80}
 -w_{\mathrm{time}}+r_{\mathrm{align},t}
 +0.3\,\mathbb{1}_{\{\text{hover region at }t\}}.
 \label{eq:navreward}
\end{align}
The stage-one configuration uses $w_{\mathrm{prog}}=0.08$, $w_{\mathrm{prog},z}=0.025$, $w_{\mathrm{ov}}=0.6$, $w_{\mathrm{ovz}}=0.5$, $w_{\mathrm{orient}}=0.25$, $w_\omega=0.05$, $w_{vh}=0.25$, and $w_{\mathrm{time}}=0.025$. The hover indicator is true only when distance, height error, horizontal and vertical speeds, and tilt are all inside their corresponding hover tolerances. The episode terminates successfully after this condition is met for 15 consecutive decisions, and a terminal bonus of 30 is then applied. A ground collision during the navigation stage or departure from the workspace incurs a terminal penalty of 30.

\subsection{Stage-two and stage-three landing rewards}

Stages two and three share a landing-oriented reward function, while differing in the relative importance assigned to speed regulation, horizontal navigation, attitude, and control effort. The reward encourages a controlled descent, progressive reduction of horizontal distance, and braking of excessive lateral motion. Let $h_t=\max(z_t-z_{\mathrm{pad}},0)$ denote the height above the landing surface. The desired vertical velocity is defined by
\begin{equation}
v_{\mathrm{des},t}
=
-\max\!\left(
V_{\mathrm{soft}},
\sqrt{2A_{\mathrm{des}}h_t}
\right),
\qquad
V_{\mathrm{soft}}=0.07,\quad
A_{\mathrm{des}}=0.0078.
\label{eq:landing_velocity_reference}
\end{equation}
This reference follows a braking-distance relationship at higher altitudes and imposes a minimum descent-speed magnitude near the surface. The speed-tracking and near-hover penalties are
\begin{equation}
\begin{aligned}
q_{s,t}
&=
\operatorname{clip}\!\left(
\frac{[v_{\mathrm{des},t}-v_{z,t}]_+}
{|v_{\mathrm{des},t}|+0.2};
0,2
\right),\\
q_{\mathrm{slow},t}
&=
\operatorname{clip}\!\left(
[v_{z,t}+0.05]_+;
0,1
\right),
\end{aligned}
\label{eq:landing_speed_terms}
\end{equation}
where $[x]_+=\max(x,0)$. The first term penalises descent that is faster than the reference profile, while the second penalises insufficient downward velocity relative to the near-hover target.

The raw stage-two and stage-three rewards combine these terms with attitude and angular-rate penalties, near-pad position shaping, horizontal-speed penalties, thrust cost, time cost, radial progress, and thrust-direction alignment:
\begin{align}
R^{(2,3)}_t
={}&
-w_s q_{s,t}
-w_{\mathrm{slow}}q_{\mathrm{slow},t}
-w_o\left(\frac{\theta_t}{90^\circ}\right)^2
\nonumber\\
&-w_{o,\mathrm{low}}e^{-h_t/80}
\min\!\left(\frac{\theta_t}{10^\circ},3\right)
-w_\omega\omega_t
\nonumber\\
&-w_{p,\mathrm{low}}e^{-h_t/100}
\min\!\left(\frac{d_t}{R_{\mathrm{pad}}},3\right)
-w_{vh}\min(v_{h,t},1)e^{-d_t/80}
\nonumber\\
&-w_{\mathrm{ov}}q_{h,t}
-w_tu_t^{\mathrm{thr}}
-w_{\mathrm{time}}
+w_{\mathrm{prog}}(d_{t-1}-d_t)
+r_{\mathrm{align},t},
\label{eq:landingreward}
\end{align}
where $d_t$ is the horizontal distance from the pad centre, $v_{h,t}$ is horizontal speed, $\theta_t$ is the tilt from vertical, $\omega_t$ is the angular-rate magnitude, $q_{h,t}$ is the horizontal overspeed measure, and $u_t^{\mathrm{thr}}\in\{0,1\}$ is the thrust command. The shaping function uses the pad radius $R_{\mathrm{pad}}=16$ simulator units. The exponential factors modulate the horizontal-speed and position penalties according to distance and altitude, while the minimum operators limit the contribution of selected terms. The progress term rewards decreases in horizontal distance, and $r_{\mathrm{align},t}$ provides directional shaping for thrust used to accelerate towards the pad or brake lateral motion.

The reward weights differ between the two stages to reflect their distinct objectives. Table~\ref{tab:landing_reward_weights} gives the values used in the implementation.

\begin{table}[htbp]
\centering
\small
\caption{Reward weights for the stage-two and stage-three landing tasks.}
\label{tab:landing_reward_weights}
\begin{tabular}{lcc}
\toprule
\textbf{Reward parameter} & \textbf{Stage 2} & \textbf{Stage 3}\\
\midrule
$w_s$                  & 0.4   & 0.4\\
$w_{\mathrm{slow}}$    & 0.5   & 0.5\\
$w_o$                  & 0.3   & 0\\
$w_{o,\mathrm{low}}$   & 0.3   & 0\\
$w_\omega$             & 0.05  & 0.05\\
$w_{p,\mathrm{low}}$   & 0.10  & 0.12\\
$w_{vh}$               & 0.2   & 0.2\\
$w_{\mathrm{ov}}$      & 0.3   & 0.6\\
$w_t$                  & 0.002 & 0.002\\
$w_{\mathrm{time}}$    & 0.02  & 0.015\\
$w_{\mathrm{prog}}$    & 0.05  & 0.08\\
$w_a$                  & 0     & 0.45\\
$w_b$                  & 0     & 0.90\\
\bottomrule
\end{tabular}
\end{table}

As shown in Table~\ref{tab:landing_reward_weights}, stage two assigns nonzero weights to both attitude penalties, supporting near-pad descent control. In stage three, these two weights are set to zero, allowing the controller to use vehicle tilt for lateral acceleration and counter-thrust during far-range navigation. Attitude nevertheless remains part of the reward's terminal shaping, as described in Section~\ref{sec:reward}.

Finally, the raw reward is scaled before being returned to PPO:
\begin{equation}
r_t=0.1R_t,
\label{eq:landing_reward_scaling}
\end{equation}
where $R_t$ includes the appropriate stage-specific step reward and any applicable terminal adjustment.

\subsection{Terminal rewards and theoretical interpretation}

At landing-stage touchdown, a softness score using its updated vertical-speed threshold $V_{z,\mathrm{lim}}=-0.2$:
\begin{equation}
 \eta_t=1-\min\!\left(\frac{|v_{z,t}|}{|V_{z,\mathrm{lim}}|},1\right),\qquad
 R_{\mathrm{success}}=40+10\eta_t+5\left(1-\frac{\theta_t}{10^\circ}\right).
\end{equation}
This bonus is added when the environment's success checker returns true. A failed touchdown receives a penalty whose components increase with excessive downward speed, tilt, pad miss distance, and horizontal speed:
\begin{align}
 P_{\mathrm{crash}}={}&15+15\min\!\left(\frac{[-v_z-|V_{z,\mathrm{lim}}|]_+}{2},1\right)
 +5\min\!\left(\frac{[\theta-10^\circ]_+}{45^\circ},1\right)\nonumber\\
 &+8\min\!\left(\frac{[d-R_{\mathrm{pad}}]_+}{100},1\right)
 +8\min\!\left(\frac{[v_h-V_{h,\mathrm{lim}}]_+}{1},1\right),
\end{align}
where $V_{h,\mathrm{lim}}=0.25$. Since $V_{z,\mathrm{lim}}=-0.2$, the vertical-speed excess term is equivalently $[-v_z+V_{z,\mathrm{lim}}]_+=[-v_z-|V_{z,\mathrm{lim}}|]_+$. It is positive only when the vertical velocity is more negative than the permitted threshold, i.e., when the downward speed exceeds the limit. The softness score uses $|v_z|$ and therefore rewards a small magnitude of vertical touchdown speed, whereas the crash term specifically penalises excessive downward speed. The two expressions have consistent signs and serve different roles. Workspace departure incurs $-30$ raw reward. In stage three, the upright-touchdown requirement is disabled, so the success label does not enforce $\theta\leq10^\circ$, even though the terminal bonus and failed-landing penalty still contain tilt-dependent terms.

The distance-difference term has a useful telescoping property. For a complete trajectory of length $T$,
\begin{equation}
 \sum_{t=1}^{T}w_{\mathrm{prog}}(d_{t-1}-d_t)=w_{\mathrm{prog}}(d_0-d_T).
 \label{eq:telescoping}
\end{equation}
Thus, without discounting, its cumulative contribution depends on net horizontal distance closed rather than on the number of intermediate steps. With PPO's $\gamma=0.995$, the discounted sum is not exactly telescoping, but the term still gives local directional feedback every decision. It is also not, in its implemented form, the strictly policy-invariant potential-based shaping term $\gamma\Phi(s_{t+1})-\Phi(s_t)$; it should be understood as an intentional shaping reward, not as a theorem guaranteeing unchanged optimal policies.

The speed-envelope penalties encode simple stopping-distance heuristics. Under constant deceleration $a>0$, the stopping-distance relation $d_{\mathrm{stop}}=v^2/(2a)$ gives $v_{\mathrm{safe}}(d)=\sqrt{2ad}$. The horizontal envelope uses $A_H=0.004$, while the vertical reference uses $A_{\mathrm{des}}=0.0078$ and a soft-speed floor $V_{\mathrm{soft}}$. These are conservative guidance references within the simulator rather than hard constraints on the reachable set: wind, attitude-dependent thrust, actuator lag, fuel consumption, and command latency all make the actual deceleration state-dependent.

The reward combines dense feedback with terminal discrimination. Distance progress and directional alignment help the policy learn approach and counter-thrust, saturated overspeed terms limit the scale of individual penalties, and the touchdown reward separates successful landings from crashes. A time penalty discourages unnecessarily long trajectories, and a thrust penalty weakly prices fuel-consuming actions. The terms are scaled by 0.1 to keep the numerical target scale manageable for the critic. However, the progress term is not clipped and the total return is not mathematically guaranteed to remain in a fixed interval; claims about bounded reward should therefore refer to the saturated components and the empirical scale observed in training, not to a strict global bound on every possible trajectory.

\section{Curriculum and PPO Training}
\label{sec:training}

The learning problem is difficult because a successful trajectory must solve several coupled subproblems under incomplete state information. The training design therefore changes the initial-state distribution and reward mode in a staged sequence, and separately controls the severity distribution of atmospheric perturbations. We describe the stage distributions, parameter transfer, weather-severity schedule, and PPO optimisation procedure. Unless stated otherwise, a training timestep denotes one environment decision; each decision advances up to three physics frames.

\subsection{Three-stage task curriculum and parameter transfer}

Let $\mathcal{M}_j=(\mathcal{S},\mathcal{A},P_j,r_j,\gamma)$ denote the stage-$j$ task, with common observation and action spaces but stage-specific initial-state distribution, reward weights, and termination logic. The curriculum is
\begin{equation}
 \mathcal{M}_1\;\longrightarrow\;\mathcal{M}_2\;\longrightarrow\;\mathcal{M}_3,
\end{equation}
where $\mathcal{M}_1$ is far approach and hover acquisition, $\mathcal{M}_2$ is near-pad soft descent, and $\mathcal{M}_3$ is integrated far-range landing. The exact distributions and budgets from the training procedure are summarised in Table~\ref{tab:training}.

In stage one, initial altitude is sampled uniformly from $[900,1300]$, vertical velocity from $[-0.8,-0.1]$ units/frame, radial horizontal distance from $[1000,1500]$ units, horizontal initial speed from $[0,0.15]$ units/frame, and initial pitch/roll offsets from $[-20^\circ,20^\circ]$. The objective is to enter a hover zone centred at altitude 300. The hover condition requires radial distance at most 10 units, altitude error at most 15 units, horizontal speed at most 0.06, absolute vertical speed at most 0.10, and tilt at most $8^\circ$, continuously for 15 decisions. This stage isolates long-range lateral navigation and the transition from acceleration to braking.

Stage two shifts to vertical control near the pad. Initial altitude is sampled from $[500,800]$, vertical velocity from $[-2.5,-0.5]$ units/frame, radial distance from $[0,15]$ units, horizontal speed from $[0,0.05]$, and initial attitude offsets from $[-15^\circ,15^\circ]$. It uses the landing reward and emphasises the vertical braking profile and terminal speed. Stage three returns to a far-range distribution with altitude $[900,1300]$, vertical velocity $[-1.5,-0.2]$, radial distance $[1000,1500]$, horizontal speed $[0,0.15]$, and pitch/roll offsets from $[-20^\circ,20^\circ]$. The final stage combines the previously learned navigation and touchdown behaviours; the stage-three configuration intentionally sets the explicit orientation-penalty weights to zero so that tilt remains available for lateral acceleration and counter-thrust.

The total training budget is $4.0\times10^6$ decisions in stage one, $4.0\times10^6$ in stage two, and $8.0\times10^6$ in stage three, totalling $16.0\times10^6$. Actor and critic parameters are transferred between stages because the observation/action dimensions and network topology remain the same. When moving from navigation reward to landing reward, the critic is reinitialised but the transferred actor is frozen for the first 100,000 decisions. Let $\theta$ denote actor parameters and $\phi$ critic parameters. During this warm-up, optimisation updates $\phi$ while constraining $\theta$ to remain at its transferred value; after the warm-up, both parameter sets update under PPO. The rationale is that the critic's targets change with the reward mode, and a poorly calibrated value estimate can create unreliable advantages for the inherited actor.

Rehearsal is used to reduce forgetting of earlier tasks. In stage two, 15\% of sampled episodes use the stage-one configuration. In stage three, 10\% of episodes use the stage-one configuration and 10\% use the stage-two configuration, with the remaining 80\% drawn from the current stage. This is episode-level task mixing, not replay of a fixed transition buffer. It seeks to preserve acquired far-navigation and descent behaviours while allowing the policy to adapt to the integrated task.

\subsection{Weather-severity curriculum}

The atmospheric level $\ell$ is sampled per episode. If the current upper cap is $c\in[0,1]$, the level distribution is a mixture
\begin{equation}
 p(\ell\mid c)=0.15\,\delta_0(\ell)+0.85\,\mathcal{U}(\ell;0,c),
 \label{eq:weatherdist}
\end{equation}
where $\delta_0$ is a point mass at calm conditions and $\mathcal{U}(0,c)$ is the uniform density on $[0,c]$. The fixed 15\% calm component is used to retain experience under nominal conditions as the rest of the distribution expands.

For a stage with budget $T_j$, maximum weather level $\ell_{\max}$, and ramp fraction $\rho=0.5$, the cap at training timestep $n$ is
\begin{equation}
 c_j(n)=\ell_{\max}\min\!\left(1,\frac{n}{\rho T_j}\right),\qquad 0\leq n\leq T_j.
 \label{eq:weathercap}
\end{equation}
Therefore, the weather support widens linearly from zero to $\ell_{\max}$ over the first half of each stage and remains at the full cap during the second half. With the default $\ell_{\max}=1$, the policy eventually sees the full range of implemented disturbances. This schedule controls the support of the training distribution; it does not guarantee monotonic improvement at every intermediate severity, and robust performance must be assessed on a held-out or separately specified evaluation distribution.

The sampling mechanism couples the magnitudes of multiple perturbations to the same $\ell$. Thus stronger wind, turbulence, gusts, sensing noise, and vehicle-model perturbations become more likely together. It supports evaluation over compound uncertainty but makes marginal disturbance effects statistically dependent. This is why the results section treats grouped rates and regression coefficients as associations, and why isolated-disturbance ablations remain an important extension.

\subsection{PPO objective, network, and optimisation schedule}

The policy is trained using Proximal Policy Optimization (PPO) as implemented by Stable-Baselines3~\cite{schulman2017ppo,raffin2021sb3}. The stochastic policy $\pi_\theta(a\mid o)$ selects the three discrete action components, while the value function $V_\phi(o)$ predicts expected discounted return. For a trajectory of rewards $r_t$, the return target is
\begin{equation}
 G_t=\sum_{k=0}^{T-t-1}\gamma^k r_{t+k},\qquad \gamma=0.995.
\end{equation}
The temporal-difference residual is $\delta_t=r_t+\gamma V_\phi(o_{t+1})-V_\phi(o_t)$ and the generalised advantage estimator (GAE) is
\begin{equation}
 \hat A_t=\sum_{l=0}^{T-t-1}(\gamma\lambda)^l\delta_{t+l},\qquad \lambda=0.95.
\end{equation}
Given the probability ratio $q_t(\theta)=\pi_\theta(a_t\mid o_t)/\pi_{\theta_{\mathrm{old}}}(a_t\mid o_t)$ and clipping parameter $\epsilon=0.2$, the clipped policy surrogate maximised by PPO is
\begin{equation}
 L^{\mathrm{clip}}(\theta)=\mathbb{E}_t\!\left[\min\!\left(q_t(\theta)\hat A_t,
 \operatorname{clip}(q_t(\theta),1-\epsilon,1+\epsilon)\hat A_t\right)\right].
\end{equation}
The clipped ratio limits the incentive for a single update to move the new action probabilities too far from the behaviour policy. The value function is trained by a squared-error loss against return targets, and an entropy term encourages exploration. In compact form, the implementation optimises a weighted combination of the negative policy surrogate, value loss, and entropy bonus; the precise loss signs follow the Stable-Baselines3 implementation.

The feature network uses two hidden layers of 128 units with $\tanh$ activations for each of the policy and value branches. Eight parallel environments are used. The number of decisions collected per environment per rollout is $n_{\mathrm{steps}}=4096/8=512$, yielding approximately 4,096 rollout samples per update. Each rollout is split into minibatches of 512 and used for 10 optimisation epochs. The initial learning rates are $3\times10^{-4}$, $1.5\times10^{-4}$, and $1\times10^{-4}$ for stages one through three. Each learning rate decays linearly to 10\% of its initial value, equivalently
\begin{equation}
 \alpha_j(p)=\alpha_{j,0}(0.1+0.9p),\qquad p\in[0,1],
\end{equation}
where $p$ is the fraction of training remaining. The entropy coefficients are $0.010$, $0.005$, and $0.003$ across the three stages. The clip range is 0.2 at every stage.

The discount factor corresponds to a characteristic geometric weighting timescale of about $1/(1-\gamma)=200$ decisions, though this is not a hard cutoff: later rewards still contribute with exponentially decreasing weight. The simulator permits up to 2,200 decisions in stage one and 3,200 decisions in stage three, so learning terminal landing behaviour remains a long-horizon credit-assignment problem. The dense progress reward, velocity references, and alignment shaping provide intermediate learning signals, while the critic estimates the downstream value of states encountered during approach and braking. The GAE parameter $\lambda$ controls the bias-variance trade-off in advantage estimation rather than changing the environment's physical horizon.

\begin{table}[H]
\centering
\small
\caption{Stage-specific configuration and common PPO settings from the training configuration.}
\label{tab:training}
\begin{tabular}{p{3.7cm}p{2.4cm}p{2.4cm}p{2.4cm}}
\toprule
Parameter & Stage 1 & Stage 2 & Stage 3\\
\midrule
Task & Far approach / hover & Soft descent & Integrated landing\\
Timesteps & $4.0\times10^6$ & $4.0\times10^6$ & $8.0\times10^6$\\
Initial horizontal radius & 1,000--1,500 & 0--15 & 1,000--1,500\\
Initial altitude & 900--1,300 & 500--800 & 900--1,300\\
Initial vertical velocity & $[-0.8,-0.1]$ & $[-2.5,-0.5]$ & $[-1.5,-0.2]$\\
Initial tilt offsets & $\pm20^\circ$ & $\pm15^\circ$ & $\pm20^\circ$\\
Initial learning rate & $3\cdot10^{-4}$ & $1.5\cdot10^{-4}$ & $1\cdot10^{-4}$\\
Entropy coefficient & 0.010 & 0.005 & 0.003\\
Rehearsal fraction & -- & 15\% Stage 1 & 10\% Stage 1 + 10\% Stage 2\\
\midrule
\multicolumn{4}{p{12.4cm}}{\textit{Common PPO settings:} $\gamma=0.995$, $\lambda=0.95$, clip $=0.2$, minibatch $=512$, epochs $=10$, parallel environments $=8$.}\\
\multicolumn{4}{p{12.4cm}}{\textit{Network:} policy and value branches each use $128\times128$ hidden units with $\tanh$ activations.}\\
\multicolumn{4}{p{12.4cm}}{\textit{Weather:} level cap ramps from 0 to 1 over the first 50\% of each stage.}\\
\bottomrule
\end{tabular}
\end{table}

\subsection{Training dynamics and interpretation of the learning curves}
\label{subsec:training_curves}

Training behaviour is analysed using episode-level reward and episode-length curves for each curriculum stage, together with the critic's explained variance. The episode-level plots show both individual outcomes and a rolling mean, allowing the central trend to be considered alongside variability induced by initial states and stochastic disturbances. Explained variance provides a complementary measure of how much return-target variation is captured by the value function. These diagnostics are interpreted jointly: reward and episode length describe task-level learning, whereas explained variance describes critic fit and is not itself a measure of landing success.

\begin{figure}[!htbp]
\centering
\begin{subfigure}[t]{0.49\linewidth}
 \centering
 \includegraphics[width=\linewidth]{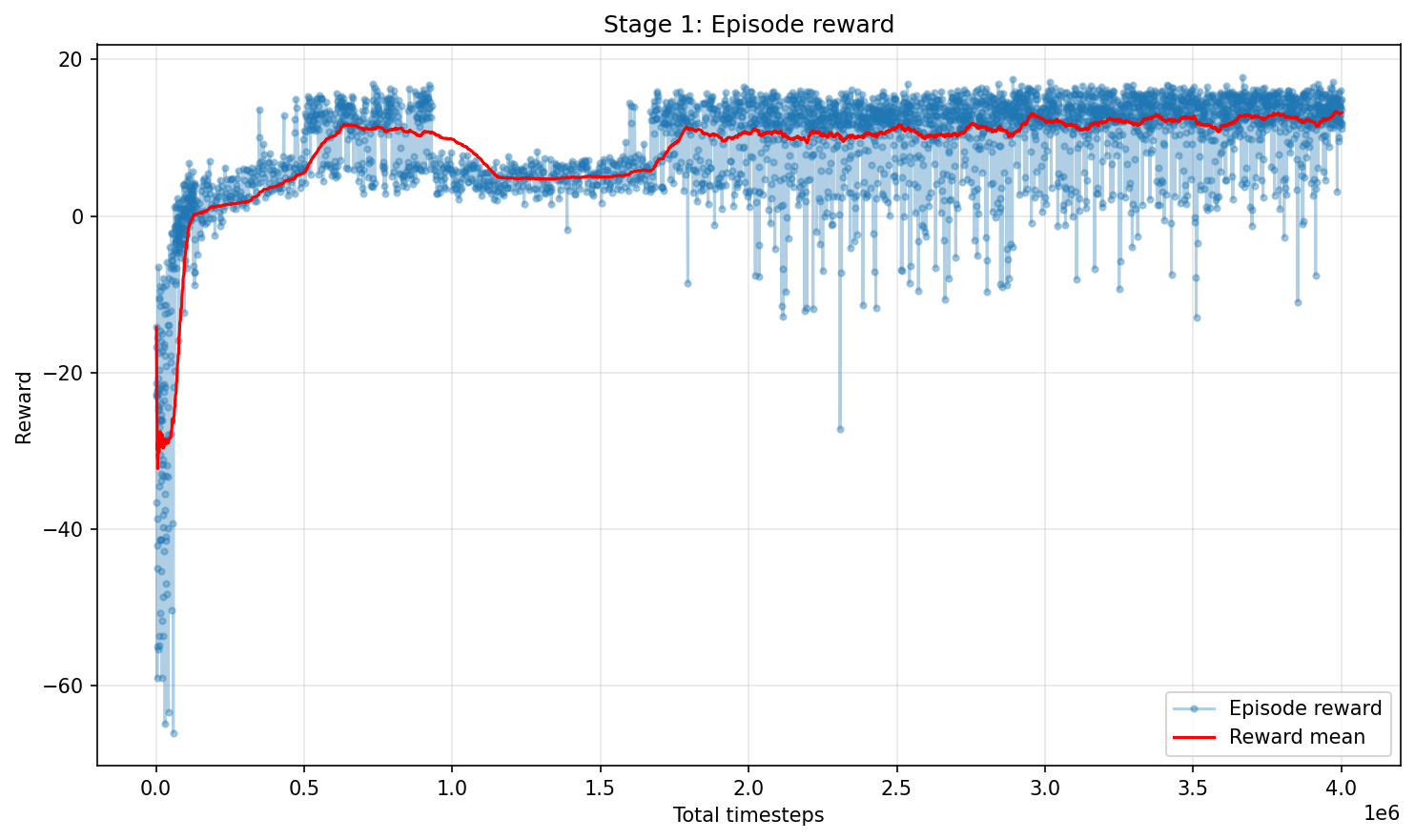}
 \caption{Episode reward}
\end{subfigure}\hfill
\begin{subfigure}[t]{0.49\linewidth}
 \centering
 \includegraphics[width=\linewidth]{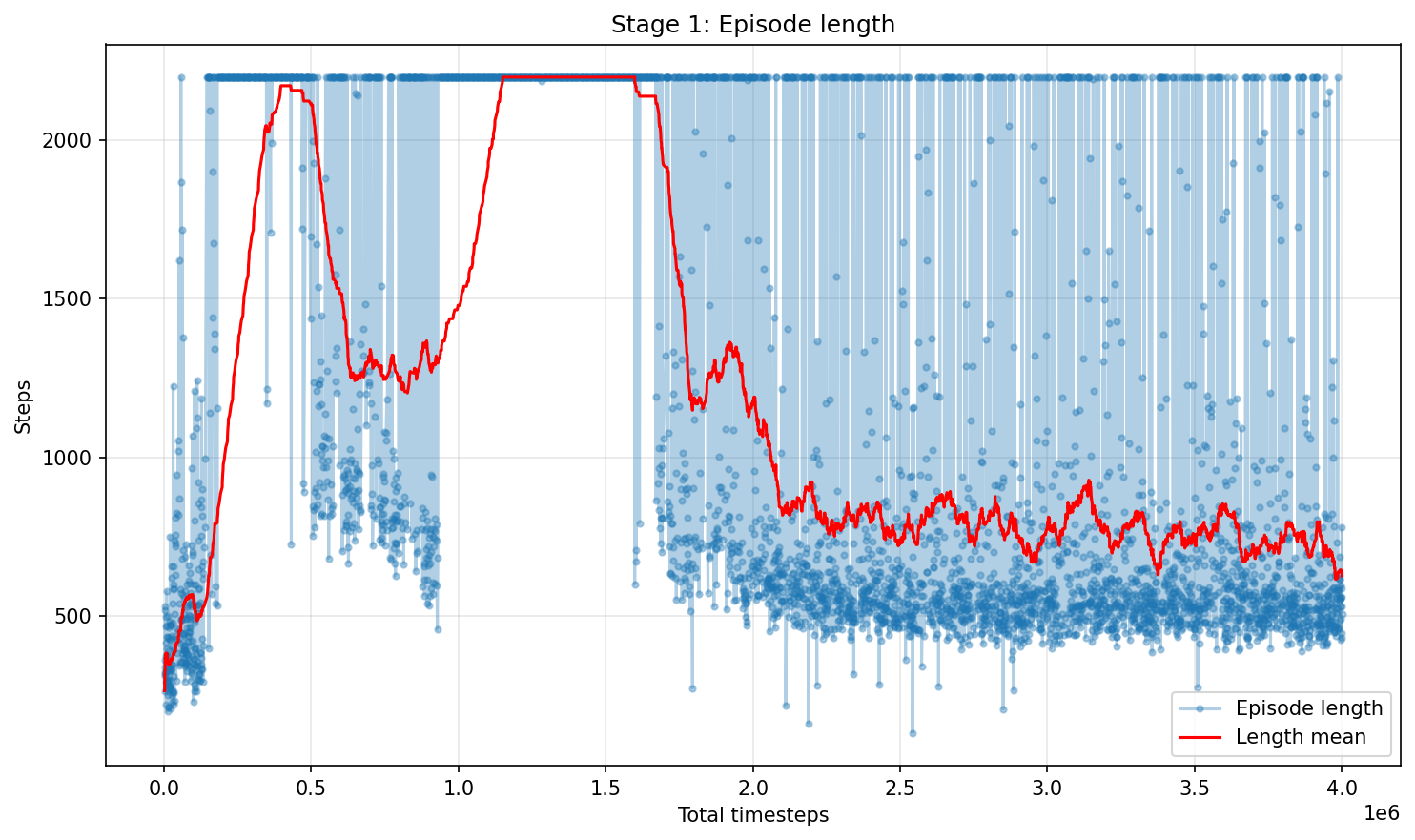}
 \caption{Episode length}
\end{subfigure}
\caption{Stage 1 training behaviour. Left: individual episode returns and their rolling mean. Right: episode lengths in agent decisions and their rolling mean. Rewards improve rapidly early in training and remain mostly positive later, while the mean episode length eventually falls from prolonged capped episodes to a shorter range.}
\label{fig:stage1_training_curves}
\end{figure}

\paragraph{Stage 1: far approach and hover acquisition.}
Figure~\ref{fig:stage1_training_curves} shows an initially sharp improvement in mean episode reward from strongly negative returns to positive values. The rolling reward then fluctuates but remains broadly positive during the latter portion of the run. Episode length initially grows and reaches the episode cap for extended periods, consistent with early exploration and difficulty satisfying the hover condition; later, the rolling mean declines to roughly 650--850 decisions. This joint pattern is consistent with the policy learning to complete the far-approach task more efficiently. It should not be read as proof that every episode reaches the hover region: the individual outcomes retain substantial spread, including occasional poor-return trajectories.

\begin{figure}[!htbp]
\centering
\begin{subfigure}[t]{0.49\linewidth}
 \centering
 \includegraphics[width=\linewidth]{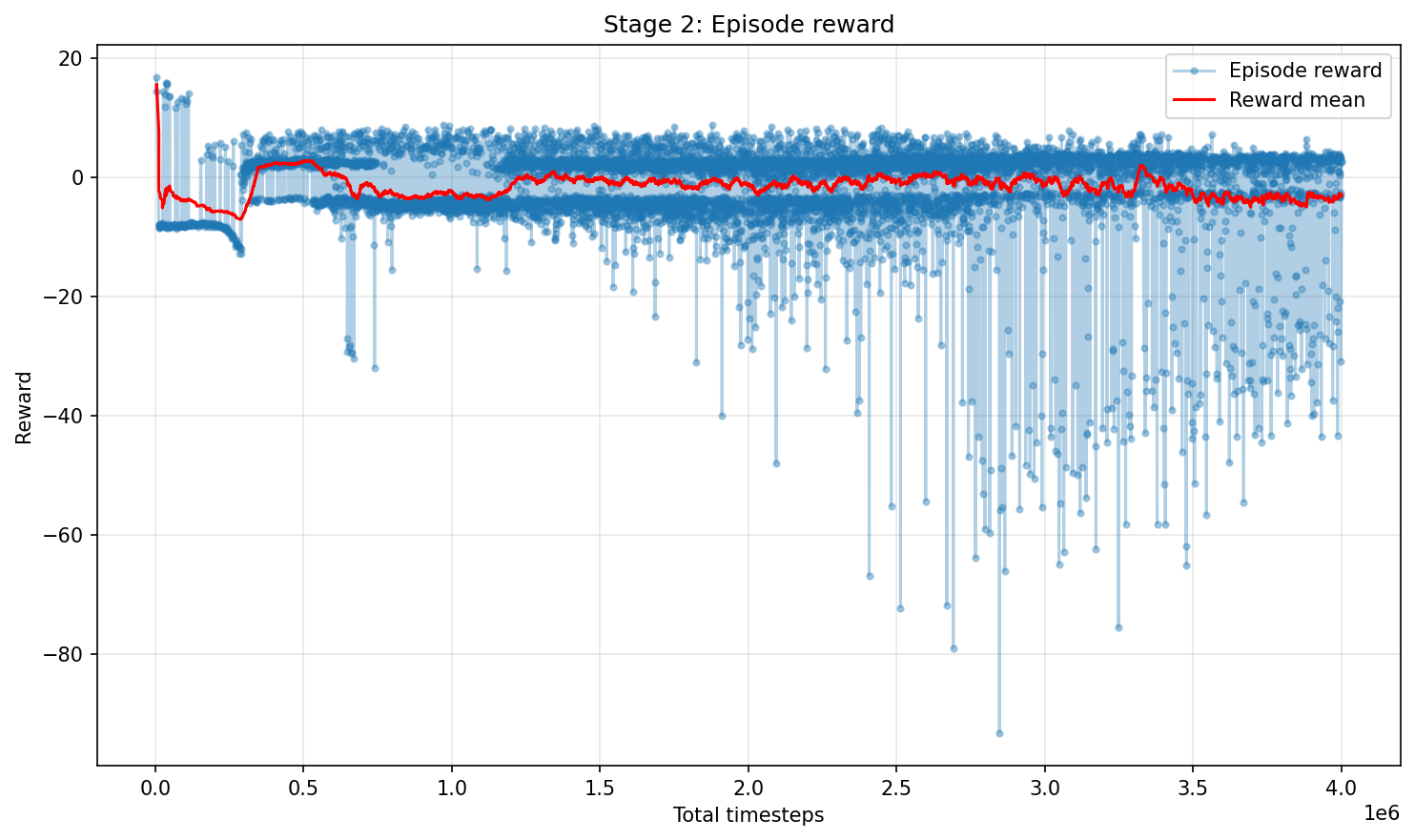}
 \caption{Episode reward}
\end{subfigure}\hfill
\begin{subfigure}[t]{0.49\linewidth}
 \centering
 \includegraphics[width=\linewidth]{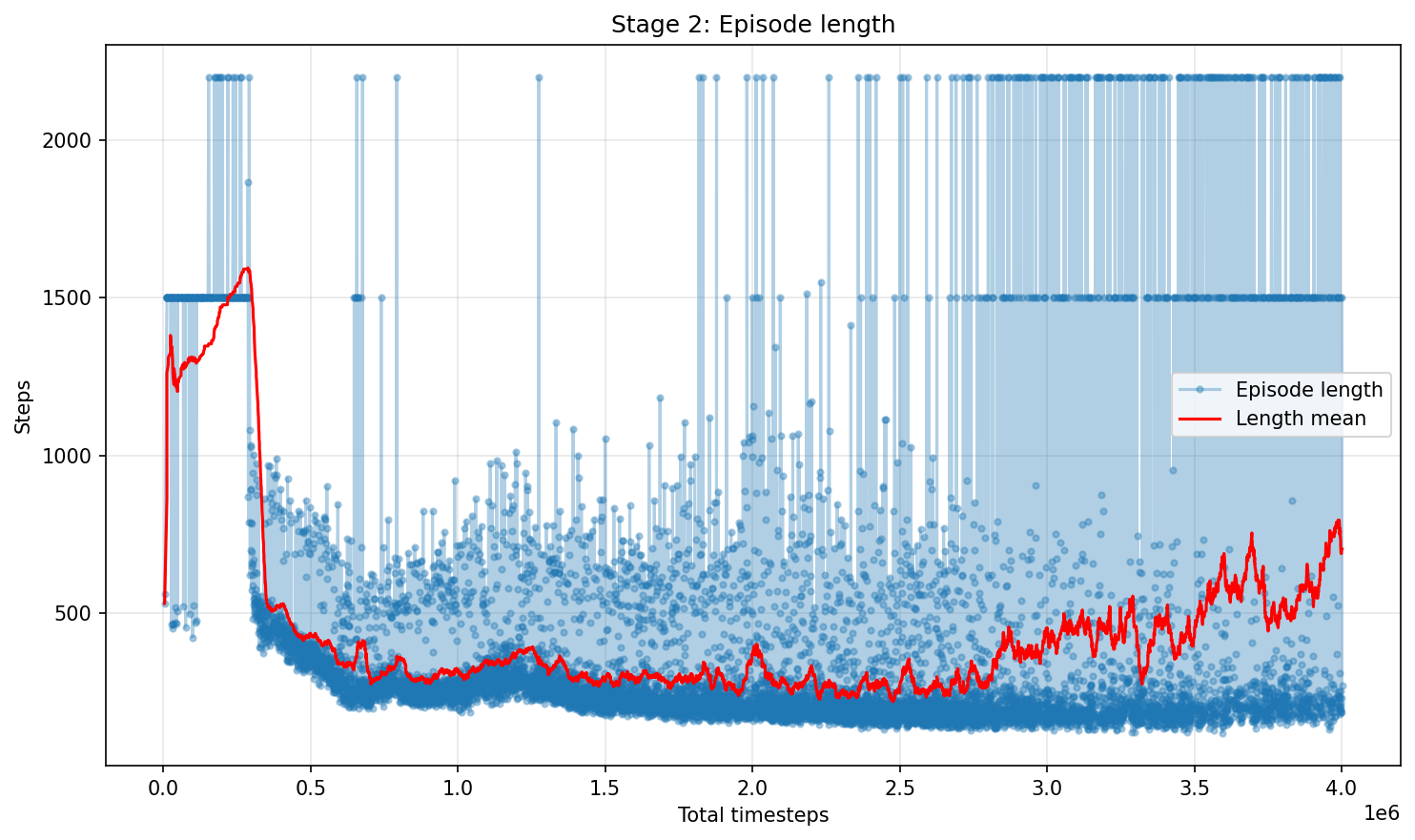}
 \caption{Episode length}
\end{subfigure}
\caption{Stage 2 training behaviour. Left: individual episode returns and rolling mean. Right: individual episode lengths and rolling mean. Both statistics are non-monotonic; the late negative reward trend and increase in mean episode length warrant checkpoint-level evaluation rather than a blanket convergence claim.}
\label{fig:stage2_training_curves}
\end{figure}

\paragraph{Stage 2: near-pad soft descent.}
The Stage 2 curves in Figure~\ref{fig:stage2_training_curves} are less monotonic. The reward mean initially changes sharply, later recovers toward values around zero, and then trends negative in the latter part of the trace. Meanwhile, episode length declines substantially early on but rises again late in the plotted trace. These changes indicate that the near-pad task's training statistics are non-stationary; they do not support a claim of smooth, monotonic convergence from the curves alone. The non-monotonic pattern may reflect interactions among the task distribution, disturbance severity, and rehearsal sampling; the curves alone do not identify which mechanism is responsible. Checkpoint-wise evaluation and controlled comparisons would be required to distinguish these explanations.

\begin{figure}[!htbp]
\centering
\begin{subfigure}[t]{0.49\linewidth}
 \centering
 \includegraphics[width=\linewidth]{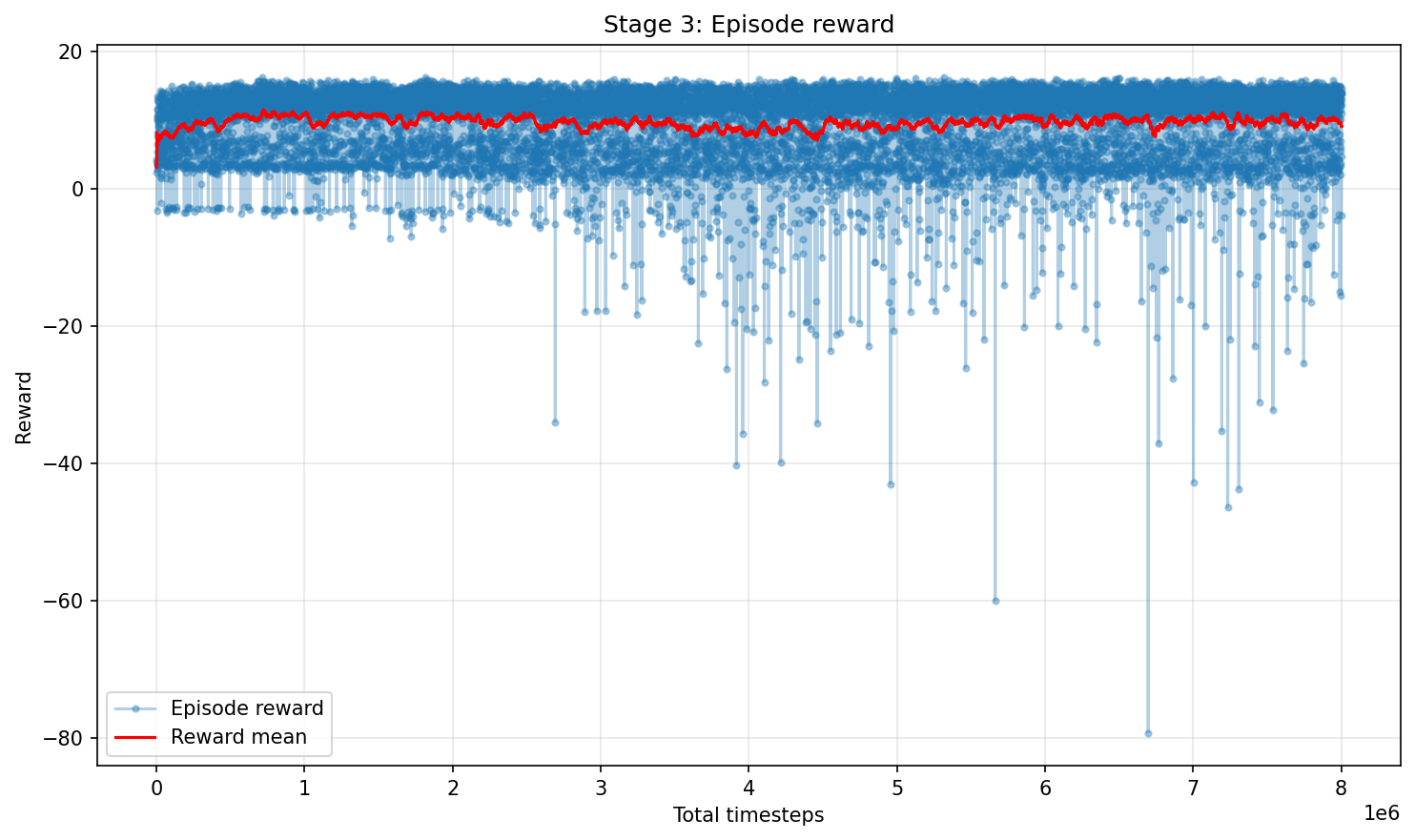}
 \caption{Episode reward}
\end{subfigure}\hfill
\begin{subfigure}[t]{0.49\linewidth}
 \centering
 \includegraphics[width=\linewidth]{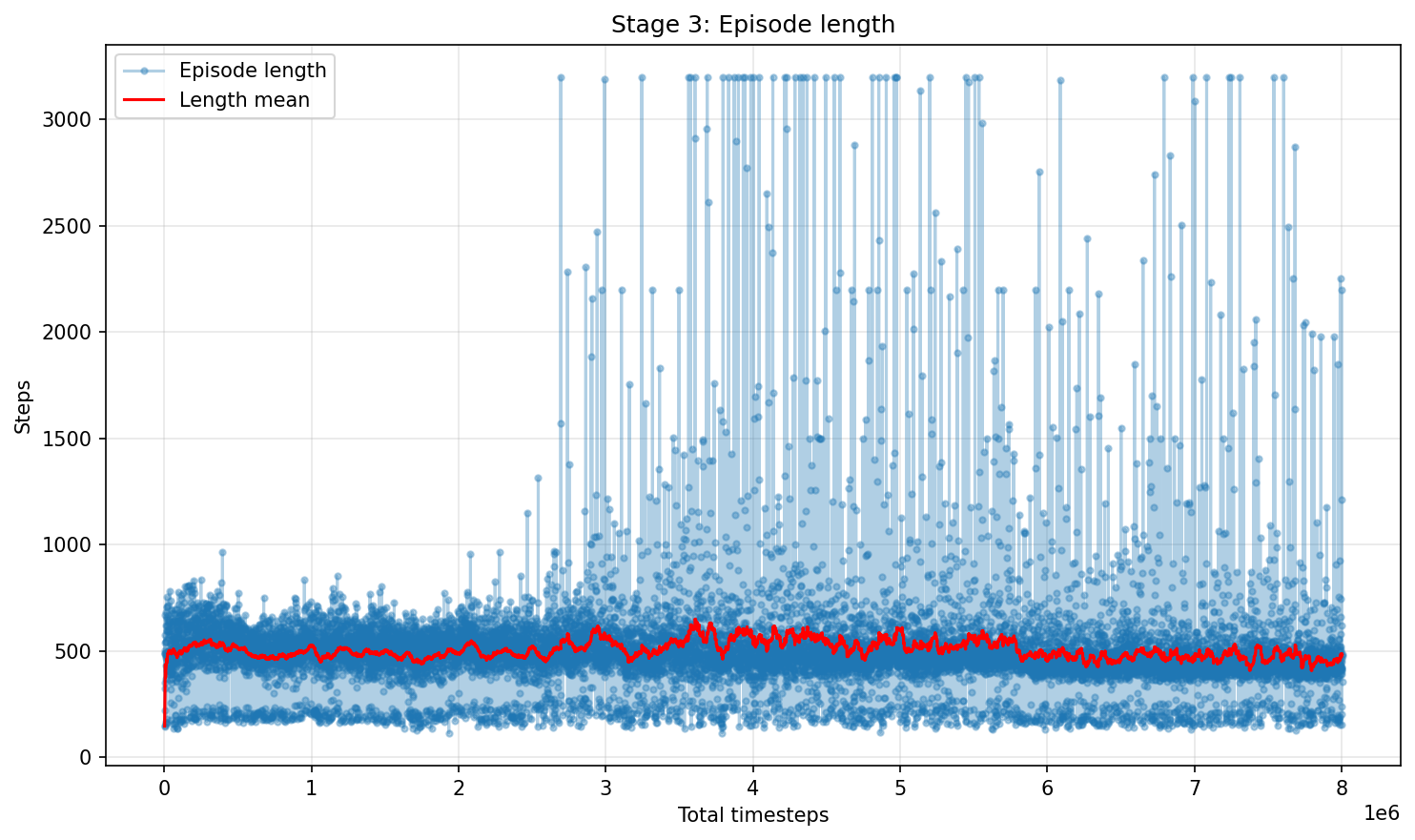}
 \caption{Episode length}
\end{subfigure}
\caption{Stage 3 training behaviour. Left: individual episode returns and rolling mean. Right: episode lengths and rolling mean. Following the initial transient, average returns remain comparatively stable and mean episode length fluctuates within a narrower band than in the early Stage 2 trace.}
\label{fig:stage3_training_curves}
\end{figure}

\paragraph{Stage 3: integrated far-range landing.}
Figure~\ref{fig:stage3_training_curves} shows the integrated task. After the initial transient, the rolling episode reward remains broadly around 9--10 for much of the run, despite individual low-return outliers. The rolling episode length fluctuates mainly around 450--550 decisions and trends somewhat lower near the end. This is consistent with a relatively stable aggregate training regime for the combined approach-and-landing task, though the individual failures and outliers show that the return is not uniform across episodes. Final success and touchdown quality must therefore be judged from the separate 100,000-episode evaluation, not from the reward curve alone.

\begin{figure}[!htbp]
\centering
\begin{subfigure}[t]{0.32\linewidth}
 \centering
 \includegraphics[width=\linewidth]{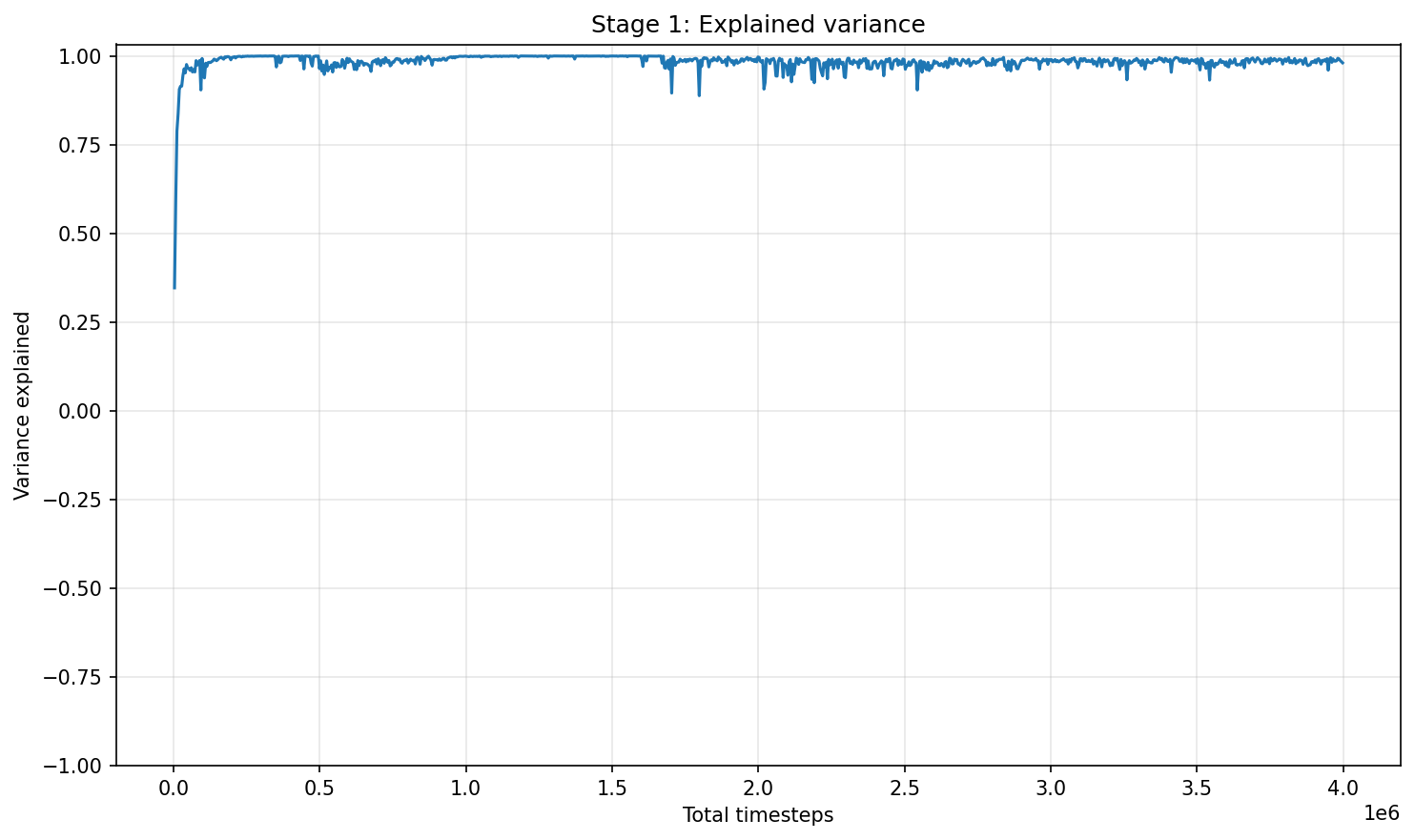}
 \caption{Stage 1}
\end{subfigure}\hfill
\begin{subfigure}[t]{0.32\linewidth}
 \centering
 \includegraphics[width=\linewidth]{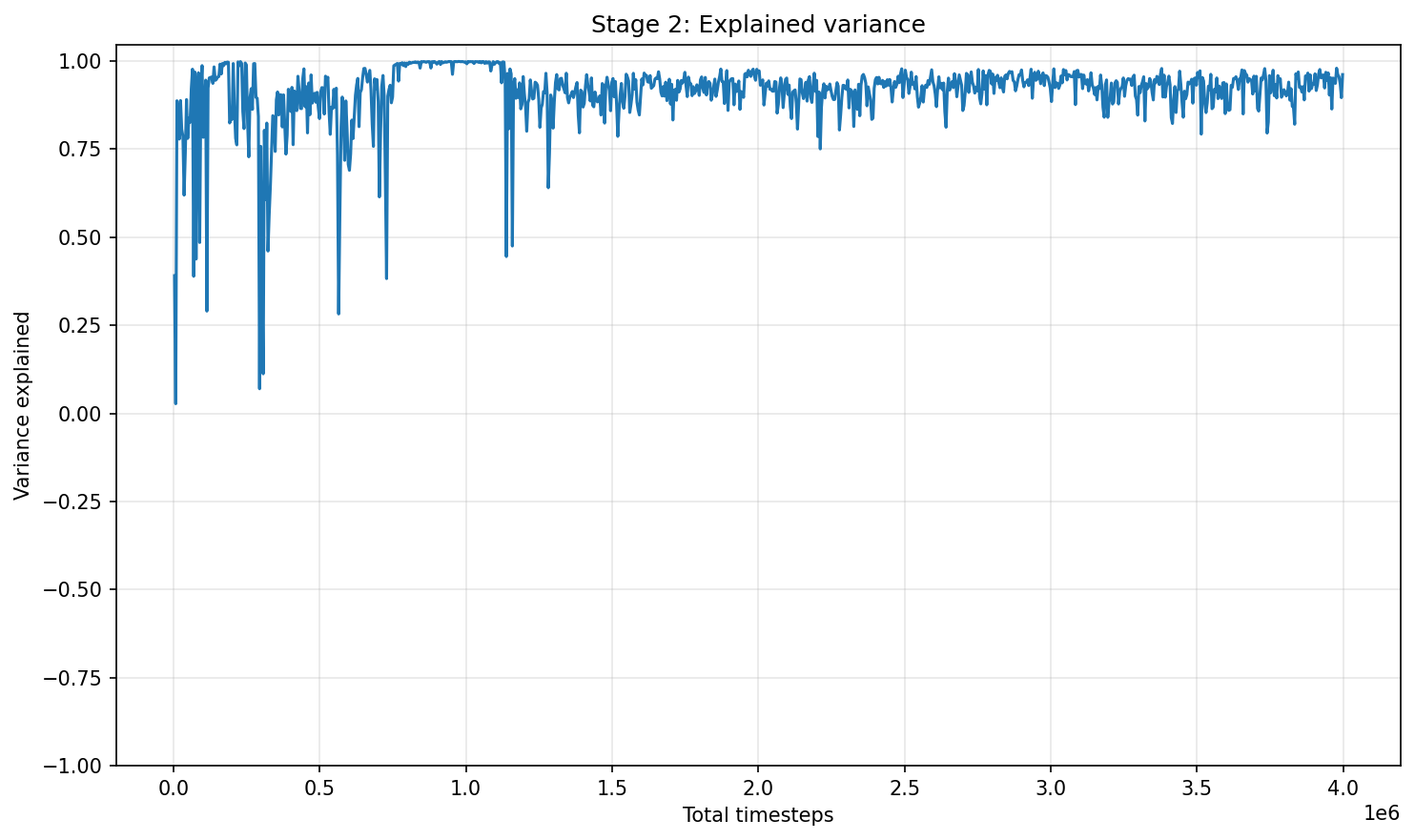}
 \caption{Stage 2}
\end{subfigure}\hfill
\begin{subfigure}[t]{0.32\linewidth}
 \centering
 \includegraphics[width=\linewidth]{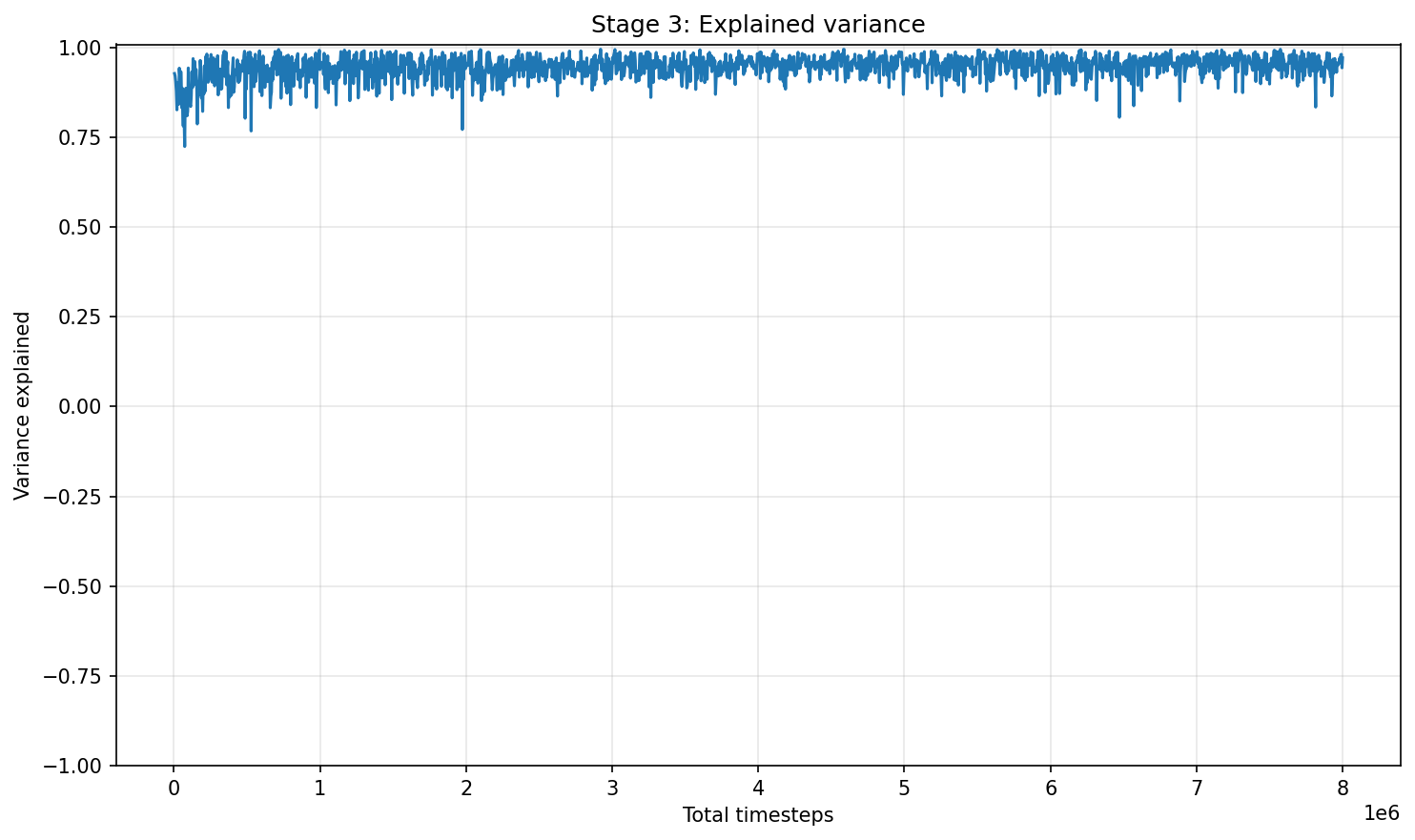}
 \caption{Stage 3}
\end{subfigure}
\caption{Critic explained variance during the three curriculum stages. Values close to one indicate that the critic explains a large fraction of return-target variance within the logged samples. Stage 2 contains more pronounced dips; in all stages the metric should be interpreted as a critic diagnostic rather than a direct measure of landing success.}
\label{fig:explained_variance}
\end{figure}

\paragraph{Critic fit across stages.}
Figure~\ref{fig:explained_variance} compares the explained-variance traces. After brief transients, the Stage 1 critic remains close to one, indicating that its value predictions explain most of the observed variance in the sampled return targets. Stage 2 exhibits more early dips and intermittent later fluctuations, which is compatible with the reward-mode change and the more variable task mixture. Stage 3 maintains a high value of explained variance for most of training, with occasional drops. Explained variance is a goodness-of-fit diagnostic for the critic on the sampled returns; it does not establish that the policy is optimal, that the learned value is calibrated out of distribution, or that the landing success rate is high. The independent evaluation in Section~\ref{sec:results} is required for those claims.

Taken together, the learning curves indicate that the far-approach and integrated stages reach comparatively stable rolling reward levels, while the Stage 2 trace is visibly less stable and deserves additional checkpoint evaluation. The plots also show why training diagnostics and test performance must be kept separate: an improving or stable average return describes the training distribution and reward function, whereas robustness is assessed by evaluating the fixed policy across the specified distribution of disturbances and initial conditions. These figures do not replace the 100,000-episode evaluation, nor do they demonstrate performance across independent training seeds.

\FloatBarrier

\section{Evaluation Protocol and Results}
\label{sec:results}

The evaluation comprises 100,000 episodes drawn across the initial-state and atmospheric-uncertainty distributions. The weather severity ranges from zero, representing the absence of stochastic weather perturbations, to the maximum modelled level. Success proportions are estimated from episode outcomes, with 95\% Wilson confidence intervals used to describe binomial sampling uncertainty for the evaluated policy. This interval reflects the finite evaluation sample and does not capture variability across independently trained policies.

A successful landing is defined by the touchdown position and translational velocity: the horizontal distance from the pad centre must not exceed 16 simulator units, the vertical velocity must satisfy $v_z\geq-0.2$ units/frame, and horizontal speed must not exceed 0.25 units/frame. The Stage 3 success criterion does not impose a separate uprightness threshold; touchdown tilt is therefore reported descriptively rather than treated as an additional condition for success. Episodes terminate following touchdown, workspace departure, or the episode time limit.

\subsection{Overall outcomes}

The model completed 96,656 successful touchdowns, 3,237 unsuccessful touchdowns, 107 timeouts, and no episodes that left the allowed workspace. Thus, the overall success rate is 96.656\%, with Wilson 95\% confidence interval $[96.543\%,96.766\%]$. The confidence interval quantifies binomial evaluation-sampling uncertainty for the evaluated policy; it does not represent variation across independently trained policies.

\begin{table}[H]
\centering
\small
\caption{Outcomes over the 100,000-episode stochastic-atmosphere evaluation.}
\label{tab:outcomes}
\begin{tabular}{lrr}
\toprule
Outcome & Episodes & Share of all episodes\\
\midrule
Successful touchdown & 96,656 & 96.656\%\\
Touchdown failing one or more criteria & 3,237 & 3.237\%\\
Timeout & 107 & 0.107\%\\
Left workspace & 0 & 0.000\%\\
\midrule
Total & 100,000 & 100.000\%\\
\bottomrule
\end{tabular}
\end{table}

\subsection{Success versus weather severity}

\begin{figure}[H]
\centering
\includegraphics[width=0.88\linewidth]{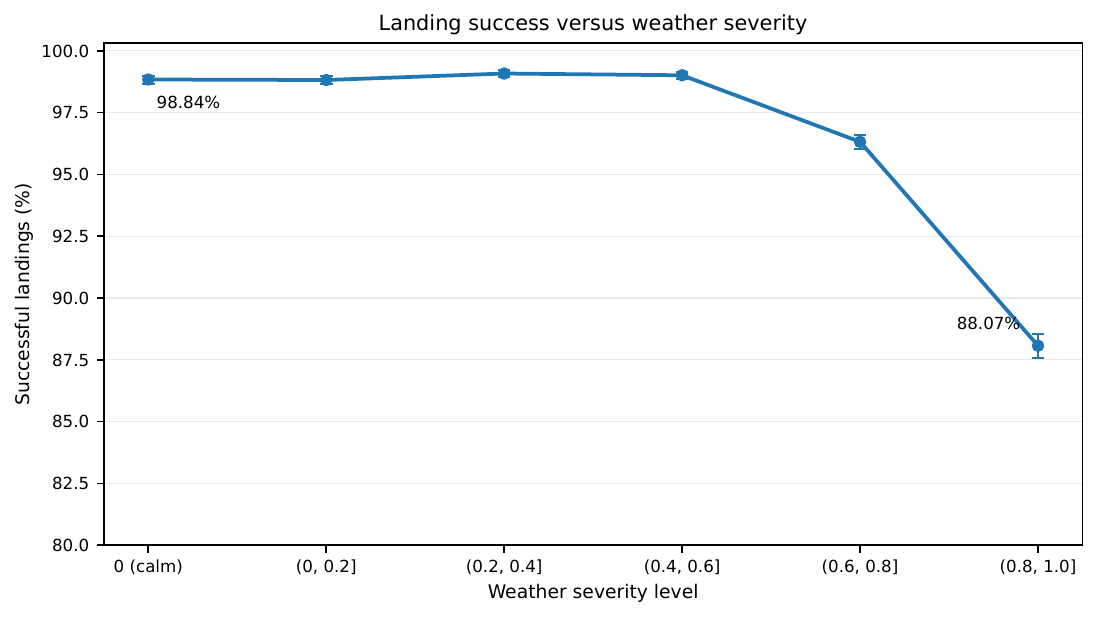}
\caption{Success rate versus weather level using the six summary bins in Table~\ref{tab:weather}. Error bars show Wilson 95\% intervals calculated from each bin's counts. The nonzero weather bins contain mixed values within their stated intervals.}
\label{fig:weather}
\end{figure}

Table~\ref{tab:weather} and Fig.~\ref{fig:weather} show that success remains near 99\% through weather level 0.6, then drops in the higher-severity bands. Success is 98.84\% for the level-zero subset, 99.08\% for $(0.2,0.4]$, 96.32\% for $(0.6,0.8]$, and 88.07\% for $(0.8,1.0]$. The $(0.6,0.8]$ band contains two timeouts, while the highest band contains 105 timeouts and 1,907 touchdown failures; no episodes are labelled as leaving the workspace. In the 0.1-wide bins, the point estimate first falls below 95\% in $(0.8,0.9]$ (91.24\%) and reaches 84.93\% in $(0.9,1.0]$.

\begin{table}[H]
\centering
\small
\caption{Outcome counts and success rate by weather severity. Rates are calculated from the episode counts.}
\label{tab:weather}
\begin{tabular}{lrrrrr}
\toprule
Weather level & $n$ & Successful & Crashed & Timeout + lost & Success\\
\midrule
0 (calm) & 15,218 & 15,041 & 177 & 0 & 98.84\%\\
$(0,0.2]$ & 17,059 & 16,857 & 202 & 0 & 98.82\%\\
$(0.2,0.4]$ & 16,936 & 16,780 & 156 & 0 & 99.08\%\\
$(0.4,0.6]$ & 16,825 & 16,658 & 167 & 0 & 99.01\%\\
$(0.6,0.8]$ & 17,097 & 16,467 & 628 & 2 & 96.32\%\\
$(0.8,1.0]$ & 16,865 & 14,853 & 1,907 & 105 & 88.07\%\\
\bottomrule
\end{tabular}
\end{table}

\subsection{Touchdown quality among successful episodes}

Touchdown-quality statistics are conditional on the 96,656 episodes labelled successful. This conditioning matters: the success rule truncates miss distance at the 16-unit pad radius and constrains touchdown speeds, so these statistics describe accepted landings rather than all attempts. Mean miss distance is $6.896\pm3.439$ units, with median 6.634 and 95th percentile 13.054 units. Mean horizontal touchdown speed is $0.0573\pm0.0310$ units/frame, and mean absolute vertical speed is $0.1085\pm0.0252$ units/frame. Fuel used is calculated as the assumed 1,000-unit initial capacity minus recorded fuel remaining; it averages $168.93\pm47.80$ units among successful episodes. Mean flight duration is $26.02\pm9.12$ seconds.

\begin{table}[H]
\centering
\small
\caption{Landing-quality statistics for the 96,656 successful episodes. Velocities are in units/frame, lengths are in simulator units, and all reported means include sample standard deviations.}
\label{tab:quality}
\begin{tabular}{lrrrr}
\toprule
Metric & Mean $\pm$ SD & Median & P95 & Criterion / note\\
\midrule
Miss distance & $6.896\pm3.439$ & 6.634 & 13.054 & $\leq16$ units\\
Horizontal speed & $0.0573\pm0.0310$ & 0.0529 & 0.1142 & $\leq0.25$\\
Vertical speed $|v_z|$ & $0.1085\pm0.0252$ & 0.1101 & 0.1470 & $v_z\geq-0.2$\\
Tilt (descriptive) & $4.932\pm2.806$ & 4.390 & 10.149 & Not in success rule\\
Fuel used & $168.93\pm47.80$ & 159.18 & 226.53 & Initial fuel: 1,000\\
Flight duration & $26.015\pm9.118$ & 24.117 & 36.517 & Descriptive (seconds)\\
\bottomrule
\end{tabular}
\end{table}

The maximum tilt among successful episodes is 50.71$^\circ$. This value is important for interpretation: touchdown at such an attitude raises serious concerns about post-contact stability for a vehicle intended to land upright. The overall success rate must therefore be understood as compliance with the three specified translational conditions, not as the probability of a physically stable upright landing. A stricter supplementary metric would add the condition $\theta\leq10^\circ$ to the existing position and velocity limits:
\begin{equation}
\hat p_{10}=\frac{1}{N}\sum_{i=1}^{N}\mathbb{1}\!\left[d_{h,i}\leq16,\;v_{z,i}\geq-0.2,\;v_{h,i}\leq0.25,\;\theta_i\leq10^\circ\right].
\label{eq:strict_upright_success}
\end{equation}
This joint success proportion cannot be determined exactly from marginal tilt summaries alone; it requires counting episodes that satisfy all four predicates simultaneously. Accordingly, no numerical estimate of $\hat p_{10}$ is reported here, and the three-condition success rate should not be interpreted as an upright-landing rate.

\subsection{Failure modes}

The 3,237 unsuccessful touchdowns violate one or more of the three landing criteria. Miss distance is the most frequently violated criterion, affecting 3,179 crashes (98.2\% of crashes), followed by excessive vertical speed in 491 crashes (15.2\%) and excessive horizontal speed in 452 crashes (14.0\%). These percentages overlap because an episode can violate more than one criterion. The mutually exclusive combinations are shown in Table~\ref{tab:failures}.

\begin{table}[H]
\centering
\small
\caption{Mutually exclusive combinations of violated touchdown criteria among the 3,237 unsuccessful touchdowns.}
\label{tab:failures}
\begin{tabular}{lrr}
\toprule
Violated criteria & Episodes & Share of crashes\\
\midrule
Missed pad only & 2,677 & 82.7\%\\
Missed pad, too fast vertically, and too fast horizontally & 366 & 11.3\%\\
Missed pad and too fast horizontally & 69 & 2.1\%\\
Missed pad and too fast vertically & 67 & 2.1\%\\
Too fast vertically only & 41 & 1.3\%\\
Too fast vertically and too fast horizontally & 17 & 0.5\%\\
\bottomrule
\end{tabular}
\end{table}

There were 107 timeouts and no workspace-exit outcomes. Fuel exhaustion was recorded in 263 episodes: 232 crashes and 31 timeouts ended with an empty tank, while no successful episode exhausted the fuel supply. In the weather table, 105 of the 107 timeouts occur in the $(0.8,1.0]$ severity band. These counts are consistent with severe disturbances increasing the risk of long or unrecoverable trajectories, though they do not isolate a causal effect.

\subsection{Sensitivity to wind, turbulence, latency, and initial state}

\begin{figure}[H]
\centering
\includegraphics[width=\linewidth]{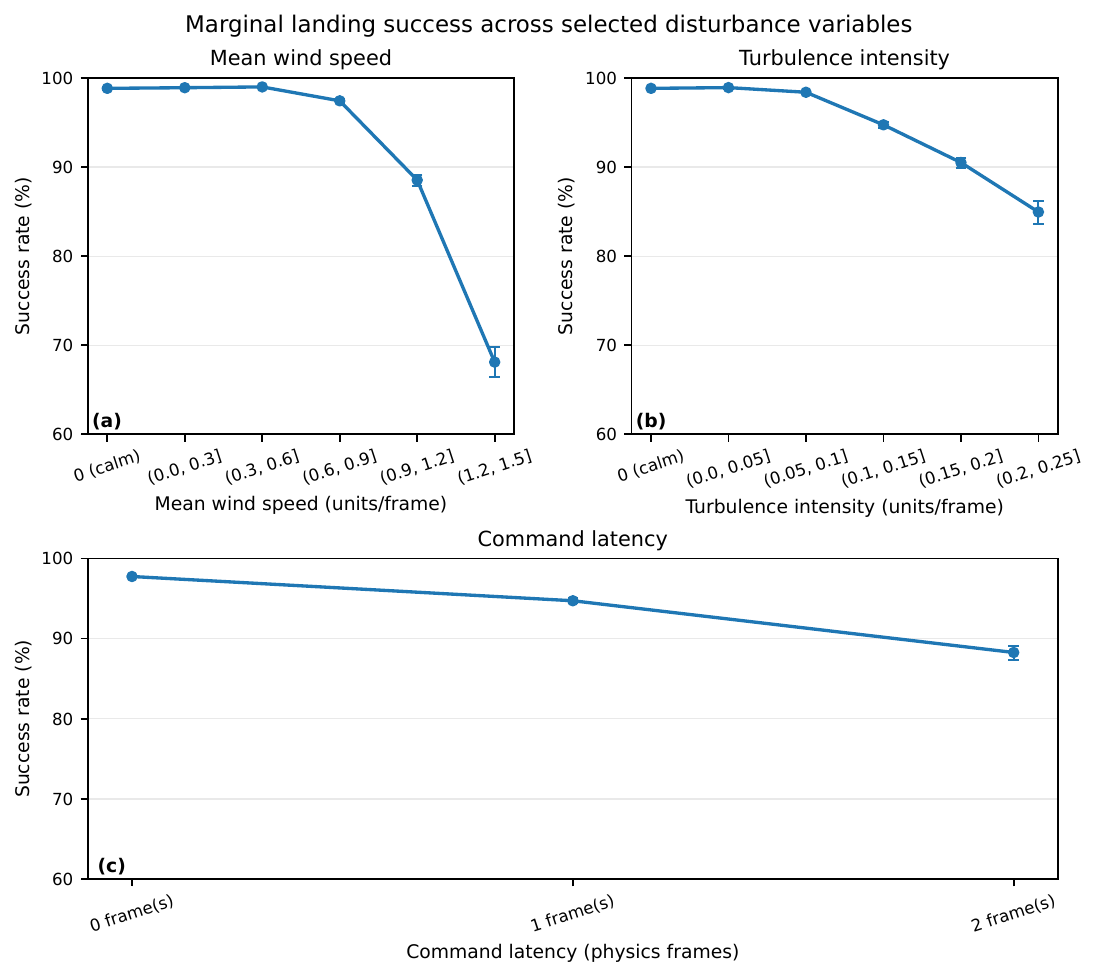}
\caption{Marginal success rates by mean wind speed, turbulence intensity, and command latency. The bins are defined over the observed ranges of the corresponding disturbance variables. The curves are descriptive and do not isolate causal effects.}
\label{fig:disturbances}
\end{figure}

Figure~\ref{fig:disturbances} and Table~\ref{tab:disturbances} show marginal success rates across selected disturbance variables. Success declines as mean wind speed and turbulence intensity increase. The rate is 98.91\% for the $(0,0.3]$ mean-wind-speed bin, 88.54\% for $(0.9,1.2]$, and 68.10\% for $(1.2,1.5]$ units/frame. For turbulence intensity, success falls from 98.92\% in $(0,0.05]$ to 84.95\% in $(0.2,0.25]$. Command latency is also associated with lower success: 97.76\% at zero frames, 94.73\% at one frame, and 88.27\% at two frames. These are marginal associations; because multiple uncertainty variables are jointly scaled by weather level, each grouping mixes other correlated effects.

\begin{table}[H]
\centering
\small
\caption{Marginal success rates for selected disturbance values.}
\label{tab:disturbances}
\begin{tabular}{llrrr}
\toprule
Factor & Bin & Episodes & Successful & Success\\
\midrule
Mean wind speed & $0$ & 15,218 & 15,041 & 98.84\%\\
& $(0,0.3]$ & 25,937 & 25,654 & 98.91\%\\
& $(0.3,0.6]$ & 25,938 & 25,678 & 99.00\%\\
& $(0.6,0.9]$ & 19,771 & 19,264 & 97.44\%\\
& $(0.9,1.2]$ & 10,145 & 8,982 & 88.54\%\\
& $(1.2,1.5]$ & 2,991 & 2,037 & 68.10\%\\
\midrule
Turbulence intensity & $0$ & 15,218 & 15,041 & 98.84\%\\
& $(0,0.05]$ & 25,936 & 25,656 & 98.92\%\\
& $(0.05,0.1]$ & 25,704 & 25,292 & 98.40\%\\
& $(0.1,0.15]$ & 19,780 & 18,740 & 94.74\%\\
& $(0.15,0.2]$ & 10,358 & 9,375 & 90.51\%\\
& $(0.2,0.25]$ & 3,004 & 2,552 & 84.95\%\\
\midrule
Command latency & $0$ frames & 74,725 & 73,048 & 97.76\%\\
& $1$ frame & 20,089 & 19,030 & 94.73\%\\
& $2$ frames & 5,186 & 4,578 & 88.27\%\\
\bottomrule
\end{tabular}
\end{table}

Initial-state grouping shows smaller but consistent patterns. Success decreases from 97.24\% in the 1,000--1,100-unit initial-distance bin to 95.83\% in the 1,400--1,500-unit bin. It rises from 95.88\% at initial heights of 900--1,000 units to 97.28\% at 1,200--1,300 units. Success is 95.89\% for initial vertical velocities between $-1.5$ and $-1.1$ units/frame, compared with 97.34\% between $-0.4$ and $-0.2$. These grouped patterns are modest relative to the high-wind degradation; as these are marginal summaries, they do not isolate initial-state effects from all co-sampled disturbance variables.

\subsection{Exploratory logistic regression}

We fit an exploratory logistic regression with standardised predictors and success as the binary outcome. Table~\ref{tab:regression} reports odds ratios per one-standard-deviation increase, ordered by the reported absolute Wald statistic. Mean wind speed is the strongest measured correlate (OR $=0.244$, 95\% CI $[0.228,0.260]$). Wind-shear exponent, initial height, and initial vertical velocity have odds ratios above one; because initial vertical velocity is negative during descent, a larger value means a less negative (slower downward) initial velocity. Initial distance has OR $=0.839$. After conditioning on the other predictors, the confidence intervals for turbulence intensity and several hardware uncertainty terms include one, so their coefficients do not provide clear evidence of independent associations in this model.

\begin{table}[H]
\centering
\small
\caption{Selected results from the in-sample logistic regression. OR is the odds ratio per one-standard-deviation increase in the named predictor.}
\label{tab:regression}
\begin{tabular}{lrrr}
\toprule
Predictor & OR & 95\% CI for OR & Reported $p$-value\\
\midrule
Mean wind speed & 0.244 & [0.228, 0.260] & $<10^{-16}$\\
Wind-shear exponent & 1.453 & [1.399, 1.508] & $<10^{-16}$\\
Initial height & 1.210 & [1.167, 1.255] & $<10^{-16}$\\
Initial vertical speed & 1.199 & [1.156, 1.243] & $<10^{-16}$\\
Initial distance & 0.839 & [0.809, 0.871] & $<10^{-16}$\\
Gust amplitude & 1.172 & [1.097, 1.252] & $2.78\times10^{-6}$\\
Command latency & 0.970 & [0.940, 1.000] & 0.0477\\
Tilt-gain error & 1.031 & [0.995, 1.069] & 0.0940\\
Initial tilt & 0.976 & [0.941, 1.012] & 0.1941\\
Turbulence intensity & 0.960 & [0.903, 1.021] & 0.1974\\
\bottomrule
\end{tabular}
\end{table}

The all-predictor model has an in-sample AUC of 0.827 and McFadden $R^2=0.205$. Because the model is fitted and assessed without a held-out split, these values describe in-sample separation and may overstate predictive performance on unseen episodes. The disturbance variables are coupled through the weather-level sampling scheme, so regression coefficients are conditional associations, not causal estimates. A chi-square test finds an association between weather bin and success ($\chi^2=4918.3$, 5 degrees of freedom, $p<10^{-16}$; Cram'er's $V=0.222$); Spearman correlation between weather level and success is $\rho=-0.175$ ($p<10^{-16}$).

\section{Discussion}

The evaluation shows that high aggregate landing success can coexist with a clear dependence on atmospheric severity. The policy achieves a success rate of 96.656\% across the evaluated distribution, with 98.84\% success under zero stochastic-weather severity and 88.07\% in the highest severity band. The decline is concentrated in severe compound conditions, where large wind, turbulence, sensor perturbations, and response delays can occur together. The aggregate success rate is therefore useful as a summary, but the severity-conditioned results provide a more informative account of the operating region in which the policy remains reliable.



The reward structure offers a plausible mechanism for learning the multi-phase trajectory. The radial progress term supplies immediate feedback on net approach, while the horizontal braking envelope discourages high lateral speed when the remaining distance is insufficient for safe stopping. Directional thrust alignment encourages the vehicle to point toward the pad while lateral acceleration is useful and against its velocity during braking. The landing reward adds vertical-speed tracking, near-ground distance shaping, time and thrust costs, and terminal feedback. The progress difference telescopes to the net distance closed in an undiscounted episode; under the discounted PPO objective it is not strictly policy-invariant potential-based shaping, but still provides a local directional signal. Because no reward-term ablation is performed, the individual causal contribution of each component cannot be determined from the final success rate alone.

The disturbance-stratified analysis indicates that mean wind speed is particularly strongly associated with landing success. Success also decreases across higher turbulence and command-latency groups, while initial displacement, altitude, and descent velocity exhibit smaller marginal associations. These relationships should be interpreted jointly rather than as isolated causal effects. The disturbance generator links several uncertainty magnitudes to the same weather-severity parameter; consequently, strong wind tends to coincide with changes in turbulence, sensor error, propulsion, and control response. The exploratory logistic regression conditions on the included predictors, but correlated and jointly sampled disturbances can still complicate attribution. Controlled experiments that vary one uncertainty family at a time, followed by factorial tests of interactions, are needed to estimate individual effects more cleanly.

Touchdown statistics for successful episodes indicate that the policy generally reaches the pad with moderate miss distance and low residual translational velocity. These metrics are conditional on success: the pad-radius and speed thresholds truncate the accepted distribution, so they do not describe the tails of unsuccessful attempts. The failure analysis shows that pad miss is the most common violated criterion, while excessive vertical and horizontal speed also contribute and can occur jointly. This points to lateral braking and terminal position regulation as important targets for future improvement. Although the reward includes a thrust cost and the analysis records fuel use, the study does not compare the policy with a fuel-optimal reference or establish optimal fuel consumption.

The training diagnostics are consistent with different levels of stability across curriculum stages. Stage 1 and Stage 3 show comparatively stable rolling-reward trends, whereas Stage 2 exhibits a more non-monotonic reward and episode-length pattern. The critic's explained variance is generally high but contains intermittent dips, particularly during Stage 2. These statistics help interpret optimisation but are not substitutes for task-level evaluation: high explained variance indicates fit to sampled return targets, not policy optimality or robust landing performance. The final conclusions are therefore based on the 100,000-episode outcome analysis together with the training curves, rather than on any single diagnostic.

Several limitations delimit the scope of the findings. The aerodynamic and vehicle perturbations are parameterised engineering models, not distributions calibrated against a specific launch vehicle or flight campaign. The simulator uses simplified translational and commanded pitch/roll attitude dynamics, approximate aerodynamic torque, and discrete control commands; it is not a high-fidelity six-degree-of-freedom vehicle model. The evaluation concerns one trained policy, so the Wilson interval quantifies episode-sampling uncertainty and does not measure variability across independent training runs. The disturbance variables are coupled by their common severity parameter, and the study does not include isolated-disturbance ablations, comparisons against a flat-PPO policy with identical observations and rewards, or systematic ablations of critic warm-up and rehearsal. Moreover, the regression results are exploratory and assessed in-sample, so their predictive performance should not be generalised to unseen episodes without validation.

The most informative next steps are to evaluate independently trained seeds; conduct controlled and factorial sweeps over wind, turbulence, gusts, observation noise, thrust and drag errors, engine lag, actuator gain, and command latency; and compare the full curriculum against flat PPO under matched training budgets. Additional studies should evaluate held-out combinations and severities, assess the contribution of rehearsal and critic warm-up, and compare the simulator against a higher-fidelity model or hardware-in-the-loop testbed. Calibration against vehicle data would be necessary to argue that the tested uncertainty distributions approximate a physical flight envelope.

\section{Conclusion}

This study investigated curriculum-driven proximal policy optimization for three-dimensional far-range rocket landing under atmospheric disturbances and vehicle-model uncertainty. The simulator incorporates altitude-dependent aerodynamic drag, wind shear, correlated turbulence, discrete gusts, uncertain density and drag, thrust variation, thrust-vector misalignment, fuel-dependent acceleration, engine lag, actuator-gain error, noisy observations, and command latency. The policy operates from a normalised observation that does not expose the latent wind or all episode-level hardware parameters, making disturbance compensation dependent on feedback from incomplete measurements.

The learning framework separates far-range approach, near-pad soft descent, and integrated landing into three curriculum stages. Policy transfer, critic warm-up after reward-mode changes, rehearsal of earlier tasks, and a progressive weather-severity schedule are combined with a stage-dependent reward that includes radial progress, braking envelopes, directional thrust alignment, vertical-speed shaping, near-ground positioning, time and thrust costs, and terminal adjustments. The architecture and reward formulation provide a structured way to learn acceleration, lateral braking, descent regulation, and terminal correction within a single feedback policy.

Across 100,000 evaluation episodes, the policy achieves 96,656 successful landings, corresponding to a success rate of 96.656\% with a 95\% Wilson confidence interval of [96.543\%, 96.766\%]. Success decreases from 98.84\% under zero stochastic-weather severity to 88.07\% in the highest weather-severity band. The disturbance analysis identifies mean wind speed as the strongest measured correlate of success and shows additional associations with turbulence, command latency, and initial-state variables. These results demonstrate strong aggregate performance within the evaluated distribution while exposing a measurable loss of reliability under the strongest compound disturbances.

The findings should be interpreted in light of the model's simplified flight dynamics, uncalibrated perturbation ranges, coupled disturbance sampling, single-policy evaluation, and absence of controlled ablations. Future work should quantify variation across independent training seeds, isolate individual uncertainty sources, compare the curriculum against matched flat-PPO training, and evaluate the policy in higher-fidelity simulation and hardware-in-the-loop settings. Such experiments are required to determine the extent to which the observed robustness generalises beyond the present simulator.

\end{document}